\documentclass{article}

\usepackage[main,final]{neurips_2026}

\usepackage[utf8]{inputenc} 
\usepackage[T1]{fontenc}    
\usepackage{hyperref}       
\usepackage{url}            
\usepackage{booktabs}       
\usepackage{amsfonts}       
\usepackage{nicefrac}       
\usepackage{microtype}      
\usepackage{xcolor}         
\usepackage{multirow}
\usepackage{graphicx}
\usepackage{amsmath}
\usepackage[most]{tcolorbox}
\usepackage[table]{xcolor}

\usepackage{wrapfig}
\usepackage{booktabs}
\usepackage{tcolorbox}

\usepackage{float}

\title{G$^2$TR: Generation-Guided Visual Token Reduction for Separate-Encoder Unified Multimodal Models}

\author{
  Junxian Li$^{1}$,\enspace 
  Kai Liu$^{1}$,\enspace 
  Zizhong Ding$^{1}$,\enspace 
  Zhixin Wang$^{2}$,\enspace \\
  \textbf{Zhikai Chen}$^{2}$,\enspace 
  \textbf{Renjing Pei}$^{2}$,\enspace 
  \textbf{Yulun Zhang}$^{1}$\thanks{Corresponding author: Yulun Zhang, yulun100@gmail.com} \enspace
  \\
  \textsuperscript{1}Shanghai Jiao Tong University,\enspace
  \textsuperscript{2}Huawei Technologies Ltd
  \\
  \small{\texttt{lijunxian0531@sjtu.edu.cn}}
  \vspace{-5.mm}
}

\begin{document}

\maketitle

\begin{abstract}
  The development of separate-encoder Unified multimodal models (UMMs) comes with a rapidly growing inference cost due to dense visual token processing. In this paper, we focus on understanding-side visual token reduction for improving the efficiency of separate-encoder UMMs. While this topic has been widely studied for MLLMs, existing methods typically rely on attention scores, text-image similarity and so on, implicitly assuming that the final objective is discriminative reasoning. This assumption does not hold for UMMs, where understanding-side visual tokens must also preserve the model’s capabilities for editing images. We propose G$^2$TR, a generation-guided visual token reduction framework for separate-encoder UMMs. Our key insight is that the generation branch provides a task-agnostic signal for identifying understanding-side visual tokens that are not only semantically relevant but also important for latent-space image reconstruction and generation. G$^2$TR estimates token importance from consistency with VAE latent, performs balanced token selection, and merges redundant tokens into retained representatives to reduce information loss. The method is training-free, plug-and-play, and applied only after the understanding encoding stage, making it compatible with existing UMM inference pipelines. Experiments on image understanding and editing benchmarks show that G$^2$TR substantially reduces visual tokens and prefill computation by \textbf{1.94$\times$} while maintaining both reasoning accuracy and editing quality, outperforming baselines on almost all benchmarks. Code is at \url{https://github.com/lijunxian111/G2TR}.
\end{abstract}

\setlength{\abovedisplayskip}{2pt}   
\setlength{\belowdisplayskip}{2pt}

\section{Introduction}

Unified Multimodal Models (UMMs) integrate the capabilities of Multimodal Large Language Models (MLLMs) and image generation models like diffusion or flux models~\cite{blackforestlabs2024flux,rombach2022high}. It can understand the tasks better with rich world knowledge from MLLMs, and then overcome complex image understanding, generation and editing tasks~\cite{liang2026rover,zhaoenvisioning,pan2026wiseedit,li2026planviz}. Modern UMMs can be categorized into two kinds based on their encoders: unified-encoder ones~\cite{team2024chameleon, xieshow} (using one encoder for understanding and generation) or separate-encoder ones~\cite{deng2025bagel,wu2025janus,tian2026internvl} (using decoupled encoders for them). Prior work~\cite{wu2025janus} suggests that decoupling visual encoding into separate pathways reduces the conflict between the visual encoder’s roles in understanding and generation. Therefore, scaling up and improving capabilities of separate-encoder UMMs have been largely explored. 

Given this, a fundamental challenge in UMMs, inference efficiency, is quite emergent for utilizing them. UMMs are typically built with billions, or even tens of billions of parameters, which leads to slow inference and substantial computational cost. This makes real-world deployment difficult, especially in resource-constrained scenarios such as edge devices or latency-sensitive interactive applications. Therefore, improving the inference efficiency of UMMs is increasingly important. Existing efforts~\cite{mao2025unimod,he2025understanding} mainly focus on modifying the internal architecture of UMMs, which often introduces considerable engineering complexity. In contrast, visual token reduction offers a simple and training-free alternative yet rarely explored. This naturally motivates our research question:

\begin{tcolorbox}[colback=yellow!10, colframe=black, boxrule=0.1mm]
\vspace{-1mm}
\textit{Can we employ visual token reduction on UMMs to improve the efficiency of them?}
\vspace{-1mm}
\end{tcolorbox}

\begin{figure}[t]
    \centering
    \includegraphics[width=0.54\linewidth]{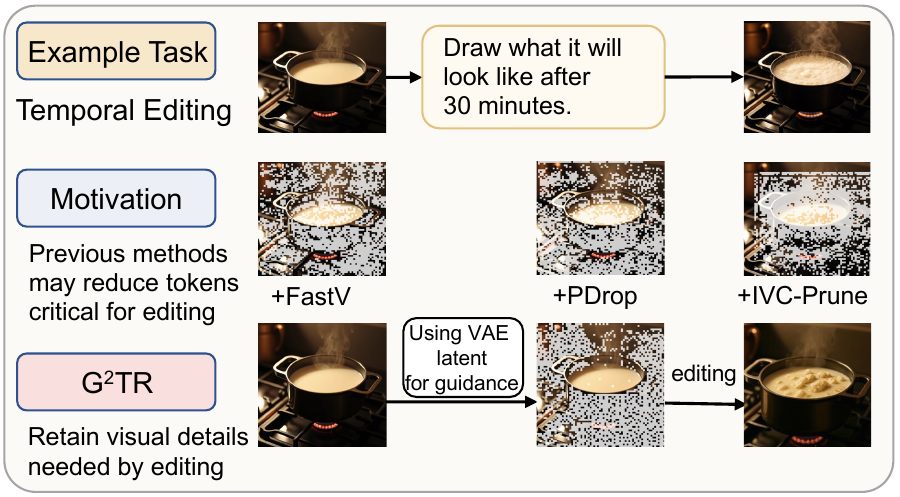}
    \hfill
    \includegraphics[width=0.42\linewidth]{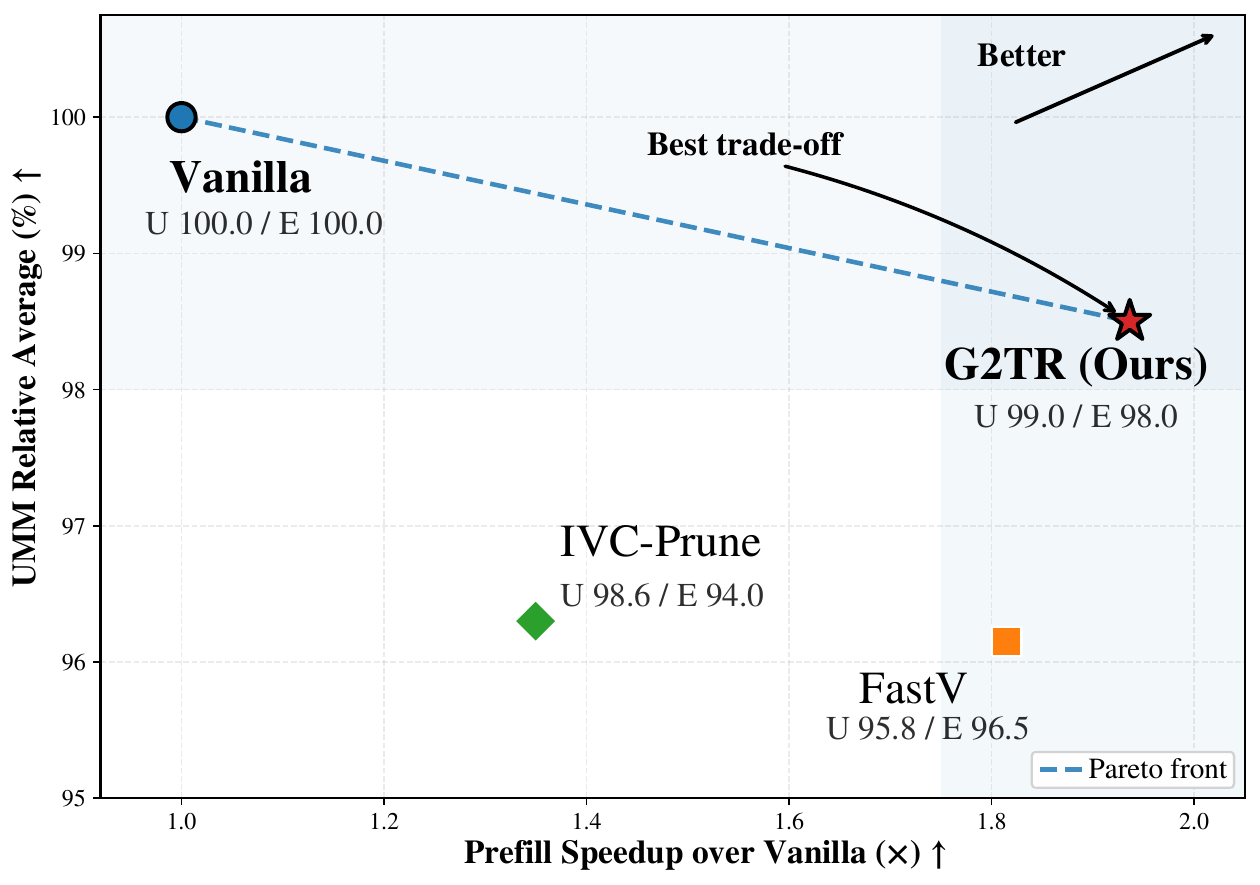}
    \vspace{-3mm}
    \caption{Intuitive view of G$^2$TR. Left: Our method preserves visual tokens crucial for image editing. Right: G$^2$TR lies on the Pareto frontier, compared with previous methods. Notably, UMM relative averages are calculated by: (understanding relative averages + editing relative averages)/2. ``U'' means understanding relative averages and ``E'' means editing relative averages (Table~\ref{tab:und} and Table~\ref{tab:quan_edit}).}
    \label{fig:teaser}
    \vspace{-7mm}
\end{figure}


Previous works~\cite{sun2026ivc,chen2024image,song2025less,xu2026rethinking,zhangvscan} have confirmed the availability of visual token reduction in MLLMs. Regarding them, we simplify the research problem in this paper to reduce understanding-side visual tokens. Thus, we mainly focus on image understanding and editing~\cite{deng2025bagel}. The mentioned methods, however, are designed for MLLMs, where visual tokens are selected for textual reasoning, e.g., by attention scores or image-text similarity. This assumption does not fully fit UMMs, since understanding-side visual tokens also condition image editing. \textbf{Tokens that seem to have lower attention or text-similarity scores may still preserve visual cues needed for high-quality editing.} Additionally, our observation is that once the pruning ratio is fixed, different pruning rules lead to only limited differences on understanding tasks. Therefore, instead of pursuing a more complex text-centric method, we focus on reducing visual tokens while preserving generation-related information.

Regarding the observations, it becomes our main goal to seek for an effective guidance signal preserving both two mentioned capabilities. Fortunately, the generation branch in UMMs naturally provides such a signal for token importance estimation. During image editing, images are compressed into latent representations that preserve generation-related information. Motivated by this, we argue that understanding-side visual tokens more aligned with these latent representations are more likely to contain visual details critical for image editing. Based on this insight, we propose \textbf{G}eneration-\textbf{G}uided Visual \textbf{T}oken \textbf{R}eduction (G$^2$TR), which leverages latents extracted from the VAE branch to guide visual token reduction on the understanding side.
Our pipeline consists of two stages: (1) important token selection and (2) redundant token merging. Considering efficiency, effectiveness, and deployment friendliness, we perform token reduction after the ViT encoder. This design serves as a plug-and-play module that can be easily applied to many separate-encoder UMMs. Consistent with our observations, this strategy may not introduce adverse effects on understanding performance. 

Extensive experiments are conducted on two separate-encoder UMMs, BAGEL-7B-MoT~\cite{deng2025bagel} and InternVL-U~\cite{tian2026internvl}. With only 50\% visual tokens, G$^2$TR preserves strong performance on both editing and understanding, and achieves the best relative average among token reduction methods, as Figure~\ref{fig:teaser} shows. On BAGEL-7B-MoT, it further reduces the KV cache by \textbf{1.90$\times$} and prefill FLOPs by \textbf{1.94$\times$}, without adding extra decoding overhead. Qualitative results also indicate that G$^2$TR keeps critical regions for understanding tasks and preserves local visual details needed for editing. These results suggest that G$^2$TR is a practical way to improve UMM inference efficiency while maintaining both understanding and editing capabilities. We summarize our contributions as follows:

$\bullet$ We propose \textbf{G$^2$TR}, a training-free and plug-and-play method for understanding-side visual token reduction for UMMs. To our best of knowledge, we first explore UMMs' visual token reduction. 

$\bullet$ We introduce and analyze the novel \textbf{generation-guided strategy}, which includes multi-stage importance estimation and token merging. This insight can be extended to other UMMs.

$\bullet$ Extensive experiments suggest that \textbf{G$^2$TR} can both preserve understanding and generation performance, while improve the efficiency of separate-encoder UMMs as well. 

\vspace{-2mm}
\section{Related Work}
\label{gen_inst}
\vspace{-3mm}

\noindent \textbf{UMMs \& separate-encoder UMMs.} UMMs integrate understanding capabilities of MLLMs~\cite{liu2023visual, wang2024qwen2, team2023gemini, zhu2025internvl3,zhang2026evaluating,zhang2026instruction,li2025chemvlm,li2026nora} and generation capabilities of diffusion or flow models~\cite{croitoru2023diffusion, rombach2022high, zhang2023adding, blackforestlabs2024flux}. Recent works on this topic~\cite{achiam2023gpt, xie2024show, chen2025janus, deng2025bagel, li2025dual,tian2026internvl,li2026unlocking,li2026planviz} try to develop end-to-end UMMs based on this idea, providing strong multimodal reasoning capabilities benefiting generation. The latest state-of-the-art models, like GPT-Image-1~\cite{openai2025b_gptimage1} and Gemini3-Pro-Image~\cite{comanici2025gemini} see a huge improvement in understanding detailed and complex instructions and generating images. Many recent open-source UMMs~\cite{chen2025janus, deng2025bagel, li2025dual} choose a separate-encoder architecture, and serve as the focus of this work.

\noindent \textbf{Efficient UMMs.} Currently, efficient UMM is a research field still largely underexplored. Several existing methods focus on techniques modifying architectures, like Mao et al.~\cite{mao2025unimod} proposed a mixture-of-depth method to solve tasks of different difficulties. While He et al.~\cite{he2025understanding} also utilizes depth pruning and width reduction on UMMs to obtain efficiency. Such methods work well, while often are more costly or not very easy for real-world utilization. In contrast, visual token reduction, serving as a simple training-free method, are rarely tried for UMMs. 

\noindent \textbf{Visual token reduction for multimodal models.} Visual token reduction is first developed in MLLMs. Usually it means two techniques: token merging and token pruning~\cite{shao2026survey}. Earlier methods simply utilize attention scores or image-text feature similarities as guidance to reduce visual tokens~\cite{chen2024image,song2025less}. Such guidance may fall short in visual redundancy, detailed structure, or task-specific token utility. More recent works~\cite{zhangvscan,sun2026ivc,xu2026rethinking} utilize techniques like RoPE scores, local-global attention scores and so on to address these issues. Adapting to separate-encoder UMMs, we design unique methods based on generation-branch guidance for better visual token reduction.

\vspace{-3mm}
\section{Preliminaries}
\label{sec:preliminary}
\vspace{-3mm}

\begin{figure}[t]
    \centering
    \includegraphics[width=0.48\linewidth]{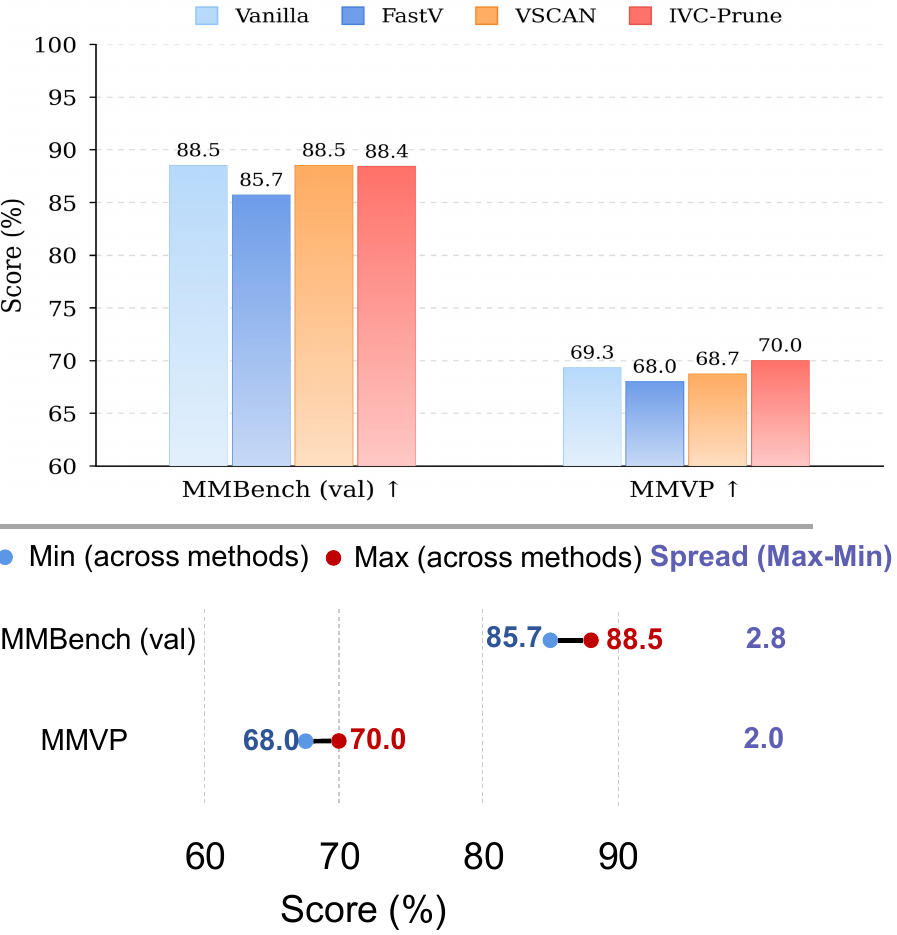}
    \hfill
    \includegraphics[width=0.47\linewidth]{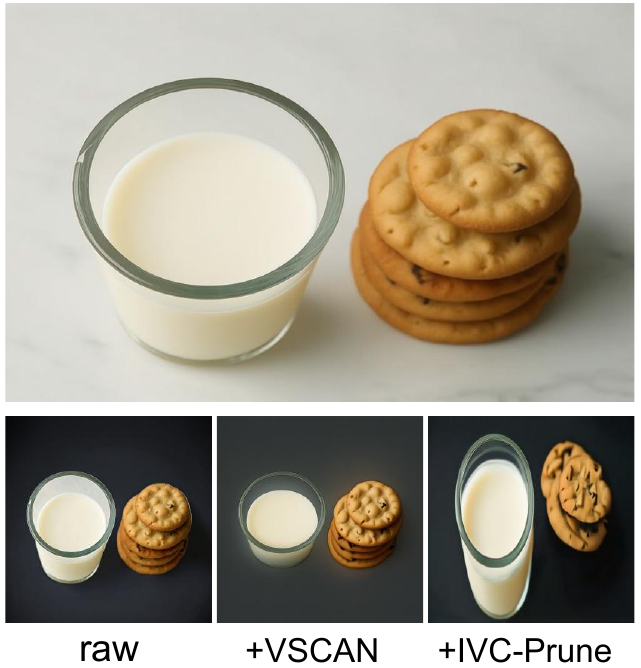}
    \vspace{-2mm}
    \caption{The two observations under this scenario. (1) Left figure indicates that limited difference exists in understanding tasks under different visual token reduction methods. (2) Right figure shows the results under prompt ``Edit this image: change the background to be darker''. Using previous visual token reduction methods cause worse editing results.}
    \label{fig:observation}
    \vspace{-5mm}
\end{figure}

\subsection{Separate-Encoder Architecture}
\label{sec:architecture}
\vspace{-2mm}
UMMs aim to support both visual understanding and visual generation within a single architecture. Recent designs~\cite{wu2025janus} suggest that these two capabilities impose different requirements on visual representations: understanding favors semantic and abstract visual features, while generation and editing requires low-level appearance and visual details. (Appendix~\ref{appendix:relationship} better shows the relationship among generation, editing, and understanding-side visual tokens.) Therefore, instead of using a single visual encoder for both tasks, a separate-encoder architecture decomposes this into an understanding branch and a generation branch. Given an input image $x$, the understanding encoder $E_{\mathrm{u}}$, typically instantiated as a ViT-style vision encoder, maps $x$ into a sequence of $N_u$ semantic visual tokens:
\begin{equation}
\mathbf U = E_{\mathrm{u}}(x) = \{\mathbf u_i\}_{i=1}^{N_u}, 
\qquad \mathbf u_i \in \mathbb{R}^{d_u},
\end{equation}
where $\mathbf U$ denotes the sequence and $\mathbf u_i$ denotes each understanding-side visual token. 

These tokens are projected into hidden state ready for inner structures $\mathbf H_{\mathrm{u}}$ by an MLP projector $P_{\mathrm{u}}$:
\begin{equation}
\mathbf H_{\mathrm{u}} = P_{\mathrm{u}}(\mathbf U) + \mathbf P_{\mathrm{u}}^{\mathrm{pos}},
\end{equation}
where $\mathbf P_{\mathrm{u}}^{\mathrm{pos}}$ denotes visual positional embeddings. This branch mainly preserves high-level semantic cues for recognition, reasoning, and instruction following. In parallel, the generation branch usually employs a VAE-style encoder $E_{\mathrm{g}}$ to obtain $N_g$ compact latent tokens:
\begin{equation}
\mathbf Z = E_{\mathrm{g}}(x) = \{\mathbf z_j\}_{j=1}^{N_g},
\qquad \mathbf z_j \in \mathbb{R}^{d_g},
\end{equation}
where $\mathbf Z$ denotes the sequence and $\mathbf z_j$ denotes each latent token. The latent tokens are mapped to the same feature hidden space through another MLP projector $P_{\mathrm{g}}$:
\begin{equation}
\mathbf H_{\mathrm{g}} = P_{\mathrm{g}}(\mathbf Z) + \mathbf P_{\mathrm{g}}^{\mathrm{pos}} + \mathbf e(t),
\end{equation}
where $\mathbf H_{\mathrm{g}}$ is the hidden state and $\mathbf e(t)$ is a timestep embedding when the model is trained with diffusion~\cite{rombach2022high} or flow-matching~\cite{blackforestlabs2024flux} objectives. The generation branch thus retains information that is essential for reconstructing or synthesizing images. Finally, the unified inner transformer layers consumes text embeddings $\mathbf H_{\mathrm{text}}$ together with tokens from either or both branches. They interact with each other through shared structures (like LLM decoders~\cite{deng2025bagel}).

This formulation highlights a useful asymmetry: ViT tokens provide semantically discriminative but often redundant visual evidence, whereas VAE latent tokens provide compact generative signals that preserve image structure. In this work, we build on this separate-encoder paradigm and study how the generation branch can guide the reduction of understanding-side visual tokens without disrupting the unified multimodal interface, thus protecting both image understanding and editing. 

\vspace{-2mm}
\subsection{Observations}
\vspace{-2mm}

\begin{table}[t]
    \centering
    \caption{Signals for guiding visual token selection of previous methods. Attention score and image-text similarity are often utilized. These designs are not suitable for directly applied in UMMs.}
    \begin{tabular}{cc}

    \toprule
       Name  & Visual Token Selection Signal \\
    \midrule
       FastV~\cite{chen2024image}  &  inner LLM attention score \\
       W-FastV~\cite{wen2025token} &  window mechanism + inner LLM attention score                        \\
       PDrop~\cite{xing2025conical} &  inner LLM attention score \\
       VSCAN~\cite{zhangvscan} &  ViT attention score + inner LLM image-text similarity \\
       IVC-Prune~\cite{sun2026ivc} & RoPE + image-text similarity \\
    \bottomrule
    \end{tabular}
    \vspace{-6mm}
    \label{tab:baselines}
\end{table}

\textit{Observation 1: different visual token reduction methods lead to limited difference on understanding capabilities.} Seeing Figure~\ref{fig:observation} left, we select BAGEL-7B-MoT~\cite{deng2025bagel} and some SOTA (state-of-the-art) text-centric visual token reduction methods~\cite{chen2024image, zhangvscan, sun2026ivc} for MLLMs. Table~\ref{tab:baselines} reports the signals for guiding important visual token selection of the SOTA methods.  ``Spread'' is defined as the maximum - minimum score on a benchmark. Different visual token reduction methods show close performance on understanding benchmarks. On MMBench, the scores range from 85.7 to 88.5, with only a 2.8 spread. On MMVP, the range is also narrow, from 68.0 to 70.0. We also find that no visual token reduction rule shows a clear and consistent advantage. This suggests that developing more complex text-centric methods for UMMs may not be very cost-effective. 

\textit{Observation 2: Visual tokens chosen by previous methods are not consistent with image editing importance.} Figure~\ref{fig:observation} right reports the visualization results of InternVL-U~\cite{tian2026internvl} (with or without token reduction methods) on image editing. We discover that, previous methods can finish the task not so well or damage local visual details, even when the instruction is simple. This shows that text-centric token importance is not enough for UMMs, and this becomes our core research problem.


\vspace{-2mm}
\section{Methodology}
\label{sec:method}
\vspace{-2mm}
\subsection{Insight and Method Overview}
\label{sec:insight}
\vspace{-2mm}
Regarding Section~\ref{sec:preliminary}, our key insight is that a ViT token well aligned with its corresponding VAE latent is more likely to carry information that should be kept for both understanding and generation. Therefore, the generation branch can serve as a natural guide for estimating understanding-side visual token importance and reducing the number of them. 

Based on this insight, we propose \textbf{G$^2$TR}, a training-free visual token reduction method for separate-encoder UMMs. 
Given an input image, we first extract full visual tokens from the understanding encoder and latent anchors from the generation-side VAE. Each visual token is then scored by its consistency with the corresponding latent anchor. 
To avoid concentrating tokens in a few salient regions, the selection is performed with balanced selection according to latent anchors. 
The removed tokens are not directly discarded. Instead, they are merged into the most similar retained tokens in feature space, which keeps the token budget low while reducing information loss, according to~\cite{zhangvscan,shaoholitom}. G$^2$TR is applied after the ViT encoder and before the inner LLM prefill. This placement keeps the original encoders and attention operators unchanged.

\begin{figure}[t]
    \centering
    \includegraphics[width=\linewidth]{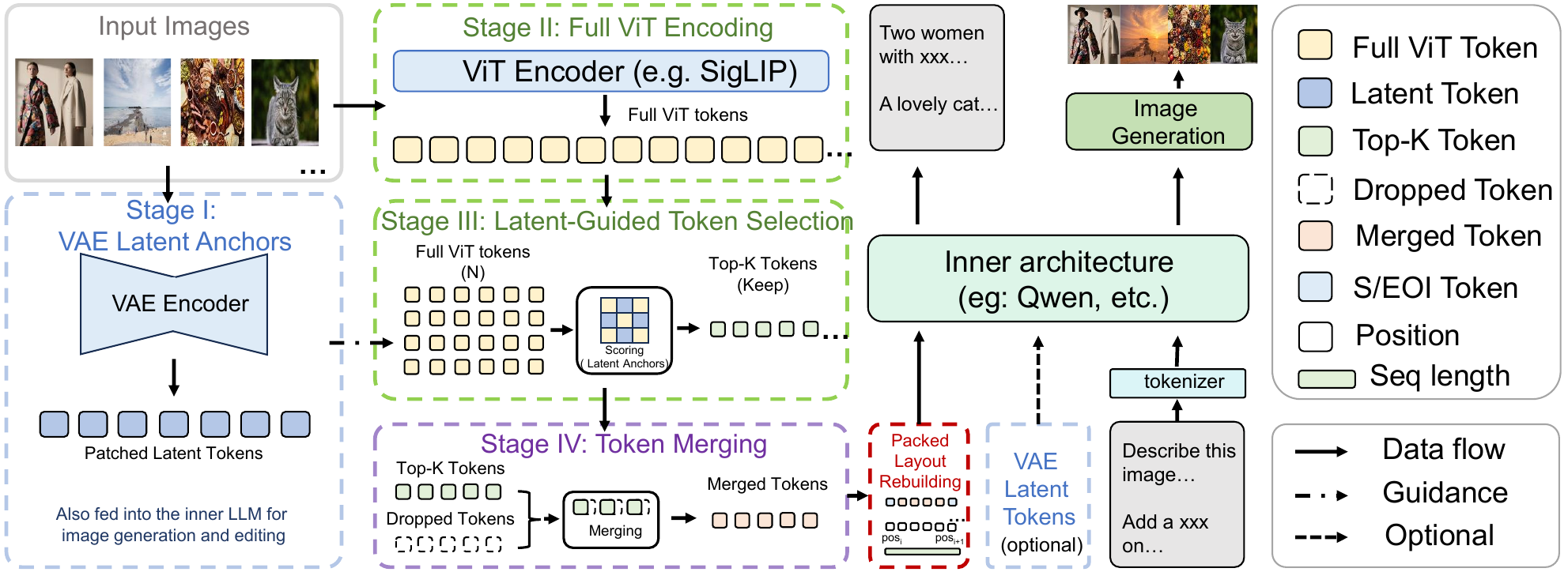}
    \vspace{-6mm}
    \caption{Method Overview. Here we focus on tasks which understanding-side visual tokens are necessary. S/EOI means ``start of image'' and ``end of image''. The whole visual token reduction process consists of four stages: firstly the VAE generates latent tokens (also serving as guidance) and the ViT generate full visual tokens; then the top-K important visual tokens are selected based on the guidance strategy; finally, the dropped visual tokens are merged onto nearest important tokens. }
    \label{fig:method}
    \vspace{-5mm}
\end{figure} 

\vspace{-2mm}
\subsection{Latent-Guided Importance Estimation}
\label{sec:importance_selection}
\vspace{-2mm}
See Figure~\ref{fig:method}, Stage III. We use the notations following Sec.~\ref{sec:preliminary}. 
Before scoring, we map them into the same hidden dimension. 
For simplicity, we still use $\mathbf{u}_i$ and $\mathbf{z}_j$ to denote the aligned features.

The sequences of ViT tokens and VAE latent tokens can then be mapped into two 2D grids. For further actions, the VAE latent tokens are specially processed. An operation similar to mean pooling is used, where four adjacent VAE latent tokens on the VAE grid are pooled into one latent anchor. Denote sizes of the ViT grids and the processed VAE grids as
$H_u \times W_u$ and $H_g \times W_g$.
For a visual token $\mathbf{u}_i$ located at $(p_i,q_i)$ on the ViT grid, we assign it to the corresponding latent anchor by
\begin{equation}
c(i)=
\left(
\left\lfloor \frac{p_i H_g}{H_u} \right\rfloor,
\left\lfloor \frac{q_i W_g}{W_u} \right\rfloor
\right),
\end{equation}
where $\lfloor \rfloor$ denotes the flooring option. We then compute its importance score as the cosine similarity to the matched VAE latent $\mathbf{z}_{c(i)}$:
\begin{equation}
s_i =
\cos(\mathbf{u}_i,\mathbf{z}_{c(i)})
=
\frac{\mathbf{u}_i^\top \mathbf{z}_{c(i)}}
{\|\mathbf{u}_i\|_2\|\mathbf{z}_{c(i)}\|_2}.
\end{equation}
A larger score means that the visual token is more consistent with the local generative latent. 
Such tokens are more likely to preserve visual factors that matter for image editing.

Given a keep ratio $\rho$, the final token budget is
\begin{equation}
K = \min\left(N_u, \max(K_{\min}, \lfloor \rho N_u \rceil)\right),
\end{equation}
where $K_{\min}$ is the minimum number of retained tokens and $\lfloor \rceil$ denotes rounding to the nearest integer.
Direct top-K selection may concentrate tokens in a few high-score regions. 
To keep the overall balance, we first select the best token inside each non-empty latent anchor:
\begin{equation}
b_{\mathbf c}=\arg\max_{i:c(i)=\mathbf c} s_i,
\qquad 
\mathbf c \in \{1,\ldots,H_g\}\times\{1,\ldots,W_g\}.
\end{equation}
This gives a candidate set $\mathcal{B}=\{b_\mathbf{c}\}$. 
We first select the top-K highest-scoring candidates from $\mathcal{B}$:
\begin{equation}
\mathcal{S}_0
=
\mathrm{TopK}\left(\mathcal{B}, \min(K,|\mathcal{B}|)\right).
\end{equation}
If $|\mathcal{S}_0|<K$, we fill the remaining budget with highest-scoring tokens from the unselected set only once:
\begin{equation}
\mathcal{S}
=
\mathcal{S}_0
\cup
\mathrm{TopK}\left(
\{1,\ldots,N_u\}\setminus \mathcal{S}_0,
K-|\mathcal{S}_0|
\right).
\end{equation}
Otherwise, we set $\mathcal{S}=\mathcal{S}_0$. 
The resulting set $\mathcal{S}$ contains the retained visual tokens.
It favors tokens aligned with generation-side latent while avoiding concentration in a few local regions.

\vspace{-2mm}
\subsection{Redundant Token Merging}
\label{sec:merge}
\vspace{-2mm}

Seeing Figure~\ref{fig:method}, Stage IV, after obtaining the retained token set $\mathcal{S}$, a simple solution is to discard all redundant tokens calculated in the selection process. 
However, direct pruning may remove fine-grained visual cues that are not selected by the importance score. 
To reduce information loss, we merge the removed tokens into the retained ones in feature space.

Let $\bar{\mathcal{S}}=\{1,\ldots,N_u\}\setminus\mathcal{S}$ denote the redundant token set. 
For each removed token $\mathbf{u}_i$ in $\bar{\mathcal{S}}$, we find its nearest retained token by cosine similarity:
\begin{equation}
n(i)=
\arg\max_{j\in\mathcal{S}}
\cos(\mathbf{u}_i,\mathbf{u}_j),
\quad i\in\bar{\mathcal{S}}.
\end{equation}
Each retained token then absorbs the removed tokens assigned to it:
\begin{equation}
\tilde{\mathbf{u}}_j
=
\frac{
\lambda \mathbf{u}_j
+
\sum_{i\in\bar{\mathcal{S}}:n(i)=j}\mathbf{u}_i
}{
\lambda+
\left|\{i\in\bar{\mathcal{S}}:n(i)=j\}\right|
},
\quad j\in\mathcal{S}.
\end{equation}
Here, $\lambda$ controls the weight of the original retained token. Following works like~\cite{zhangvscan}, we set $\lambda$ just to 1.0. Finally, the final compressed visual sequence is
\begin{equation}
\tilde{\mathbf{U}}=
\{\tilde{\mathbf{u}}_j \mid j\in\mathcal{S}\}.
\end{equation}
We keep the order of the retained indices when forming $\tilde{\mathbf{U}}$. 
This makes the output sequence shorter while preserving part of the information from removed tokens. With our design, the inner architectures of UMMs receive a much shorter sequence of understanding-side tokens.

\vspace{-2mm}
\subsection{Layout Rebuilding}
\label{sec:layout}
\vspace{-2mm}
After merging, the shortened visual tokens usually no longer match the original \texttt{<IMG\_CONTEXT>} span. 
This alignment is important for UMMs, since the inner LLM hidden states also condition image editing. 
We therefore rebuild the input before the inner LLM prefill: the original understanding-level visual tokens are replaced by the merged tokens $\tilde{\mathbf{U}}$, while text tokens and image boundary tokens are kept unchanged. 
The input ids, attention mask, and position ids are updated and padded across the batch. 
This keeps the inner UMM architecture unchanged, while letting both the inner LLM and the generation decoder work on the same shortened sequence.

\vspace{-2mm}
\subsection{Flash-Attention Compatibility}
\label{sec:flash_attn}
\vspace{-2mm}
Modern large models are often accelerated through Flash-Attention~\cite{dao2022flashattention}. Given this, a practical advantage of our G$^2$TR is its strong compatibility with this acceleration support. The implementation of existing methods, such as VScan-like~\cite{zhangvscan} variants, often need explicit attention maps to estimate token importance, or introduce pruning inside the language model. This may require extra standard attention score computation and complicate KV-cache management under Flash-Attention. This issue happens in common implementations, where Flash-Attention is used for the forward pass but attention weights are separately reconstructed by dense matmul and softmax for token scoring. In contrast, our method performs token reduction only before inner prefilling. The inner architecture therefore only receives a shorter sequence, without changing the attention operator or requesting attention weights. This plug-and-play design for separate-encoder UMMs keeps the original Flash-Attention kernel unchanged while directly reducing computational cost. 

\vspace{-2mm}
\section{Experiments}
\vspace{-2mm}
\subsection{Experiment Settings}
\label{sec:experiment_settings}
\vspace{-2mm}
The purpose of our experiments is to answer two questions: (1) can our method keep understanding and editing performance; (2) can our method really improve efficiency of UMMS. 

\textbf{Models.} We mainly choose one famous separate-encoder UMM, BAGEL-7B-MoT (BAGEL-7B)~\cite{deng2025bagel}. A very recent separate-encoder UMM, InternVL-U (4B)~\cite{tian2026internvl}, is also employed. The evaluation codes are based on their official repository. Scores of InternVL-U are reproduced on our own.

\textbf{Benchmarks.} We evaluate the performances on tasks where understanding-side visual tokens are necessary: image understanding and image editing. For image understanding, we choose MME~\cite{fumme}, MMBench~\cite{liu2024mmbench}, MMVP~\cite{tong2024eyes} and RealWorldQA (RWQA)~\cite{xai_grok1_5}. Results of BAGEL-7B on MMVP are implemented by ourselves. For image editing, we adopt GEdit-Bench (GEdit, only evaluating English queries)~\cite{liu2025step1x}, IntelligentBench (Intell.)~\cite{deng2025bagel}, and RISE-Bench (RISE)~\cite{zhaoenvisioning}. 

\textbf{Baselines.} Following IVC-prune~\cite{sun2026ivc} (ICLR'26), we choose some SOTA visual token reduction methods from top-tier venues:  FastV~\cite{chen2024image}, Window FastV (W-FastV)~\cite{wen2025token}, PDrop~\cite{xing2025conical}, VSCAN~\cite{zhangvscan} and IVC-prune~\cite{sun2026ivc} itself. We use the official implementation method and layers for visual token reduction from their official repository. ``Vanilla'' means the original performance of utilized UMMs. 

\textbf{Other settings.} We test all the models, benchmarks and baselines on one A6000 48G GPU once. The default generation hyper-parameters are chosen. Details of utilized models, benchmarks, and hyper-parameters are in Appendix~\ref{appendix:detailed_experiment}.

\begin{figure}[t]
\scriptsize
\centering
\resizebox{\linewidth}{!}{

\begin{tabular}{cccccc}
\includegraphics[width=0.15\textwidth, height=0.08\textheight]{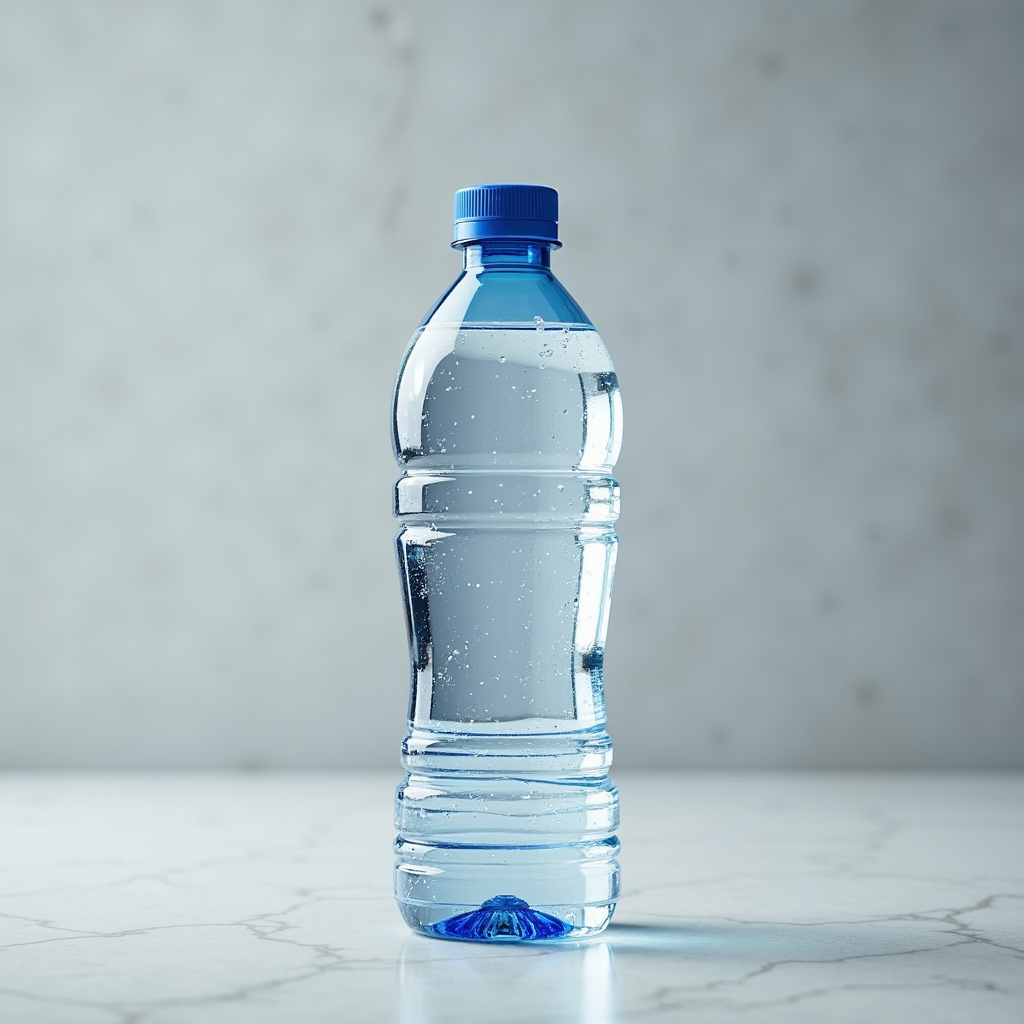} \hspace{-5mm}  &
\includegraphics[width=0.15\textwidth, height=0.08\textheight]{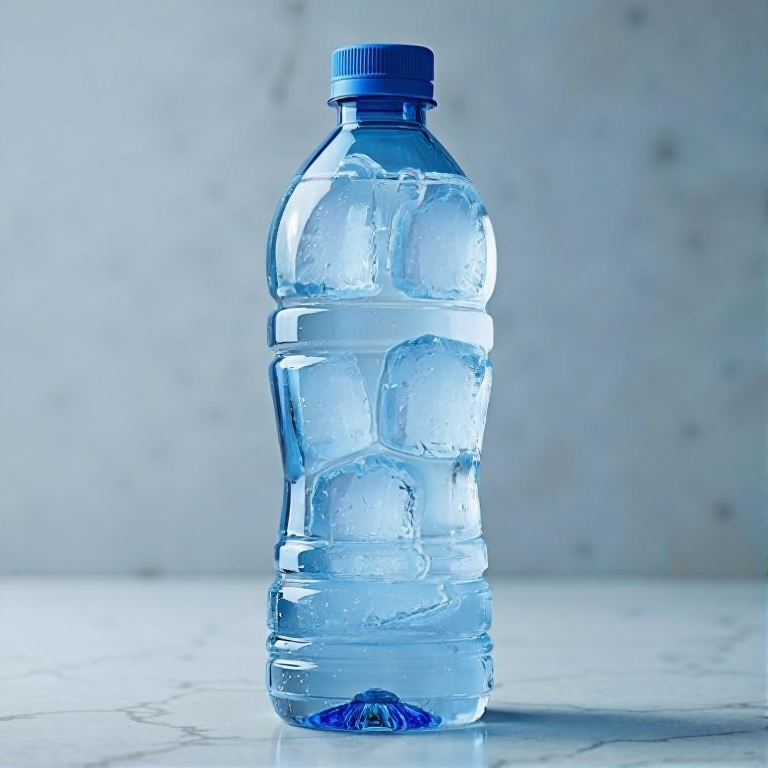} \hspace{-5mm}  &
\includegraphics[width=0.15\textwidth, height=0.08\textheight]{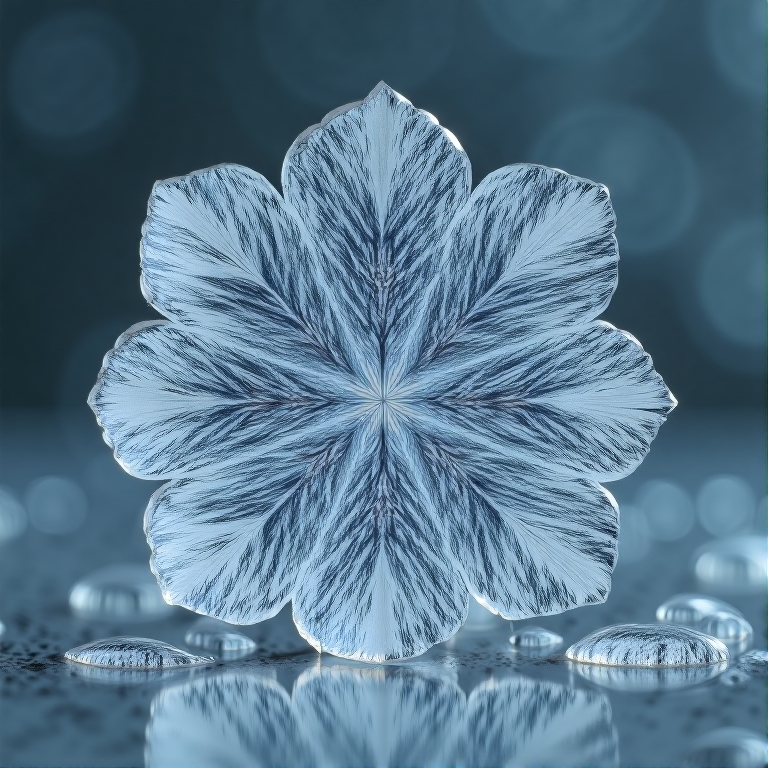} \hspace{-5mm}  &
\includegraphics[width=0.15\textwidth, height=0.08\textheight]{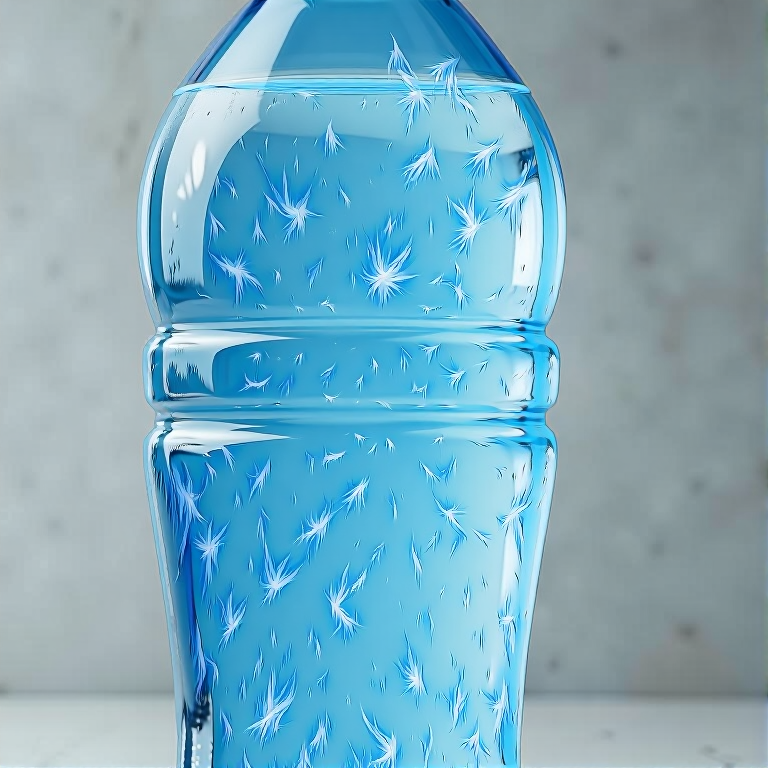} \hspace{-5mm}  &
\includegraphics[width=0.15\textwidth, height=0.08\textheight]{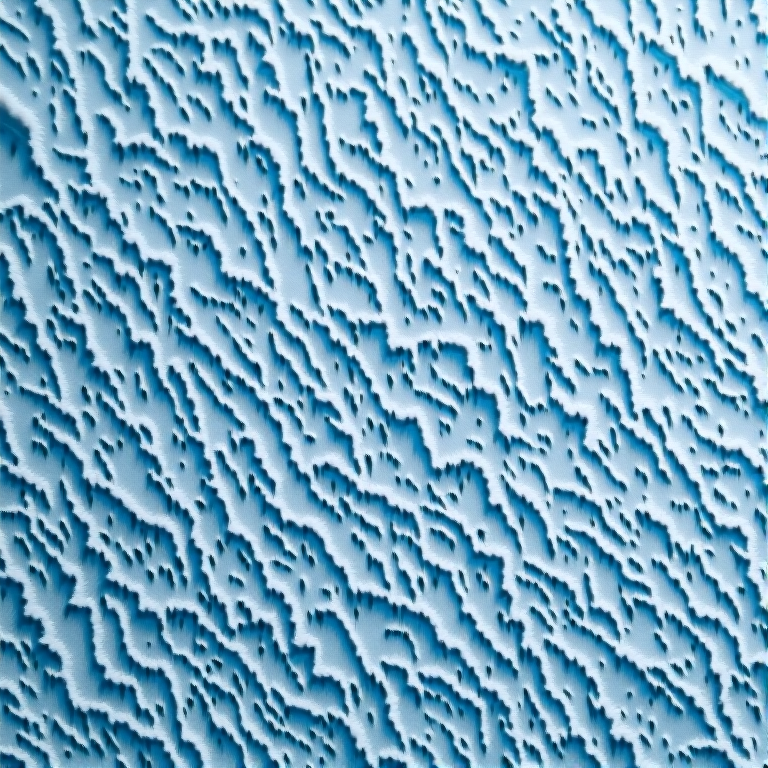} \hspace{-5mm}  &
\includegraphics[width=0.15\textwidth, height=0.08\textheight]{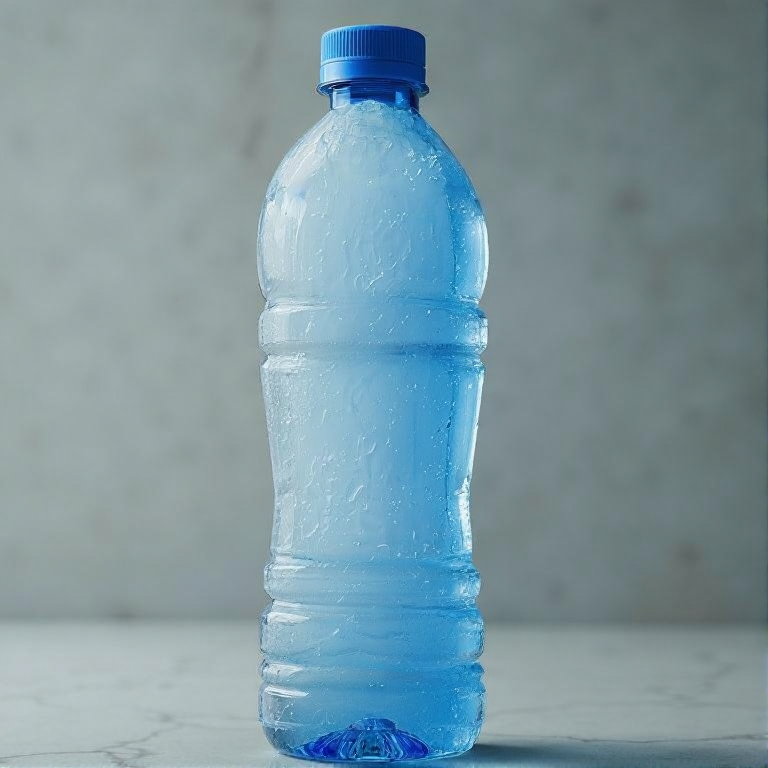} \hspace{-5mm}  
\\
\includegraphics[width=0.15\textwidth, height=0.08\textheight]{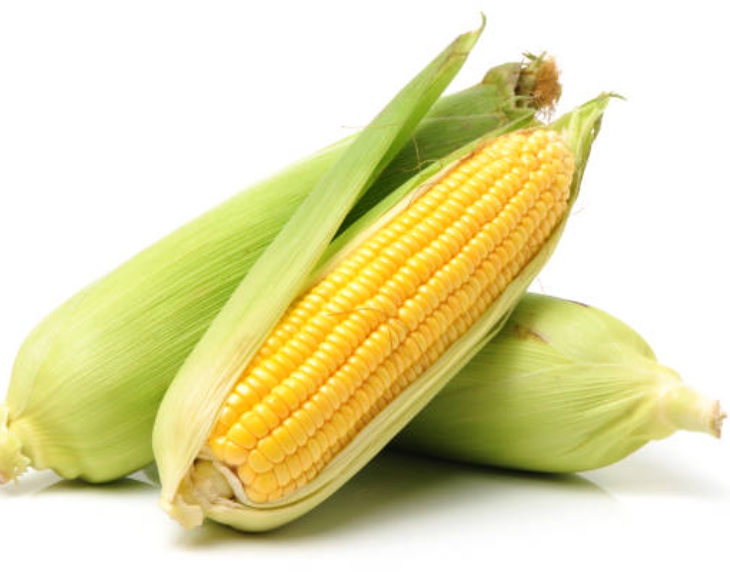} \hspace{-5mm}  &
\includegraphics[width=0.15\textwidth, height=0.08\textheight]{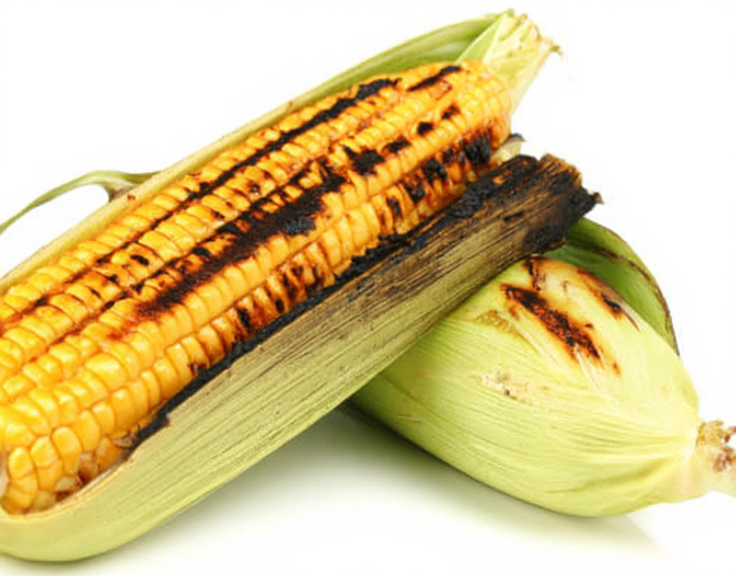} \hspace{-5mm}  &
\includegraphics[width=0.15\textwidth, height=0.08\textheight]{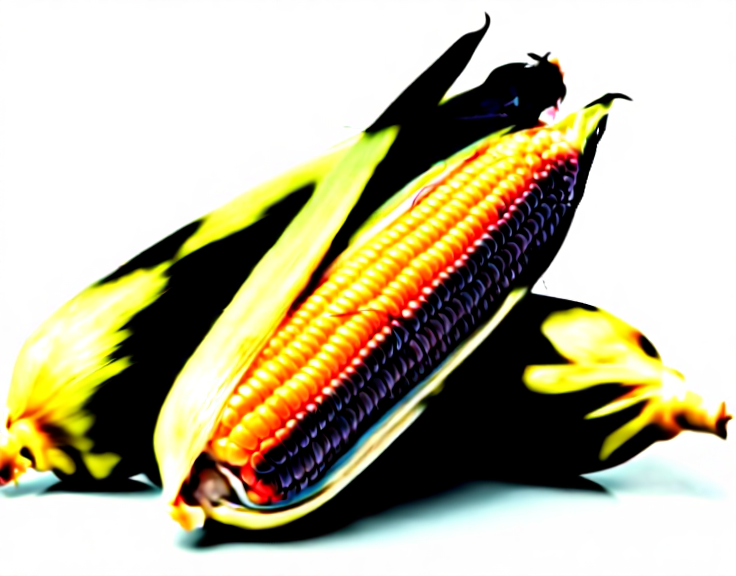} \hspace{-5mm}  &
\includegraphics[width=0.15\textwidth, height=0.08\textheight]{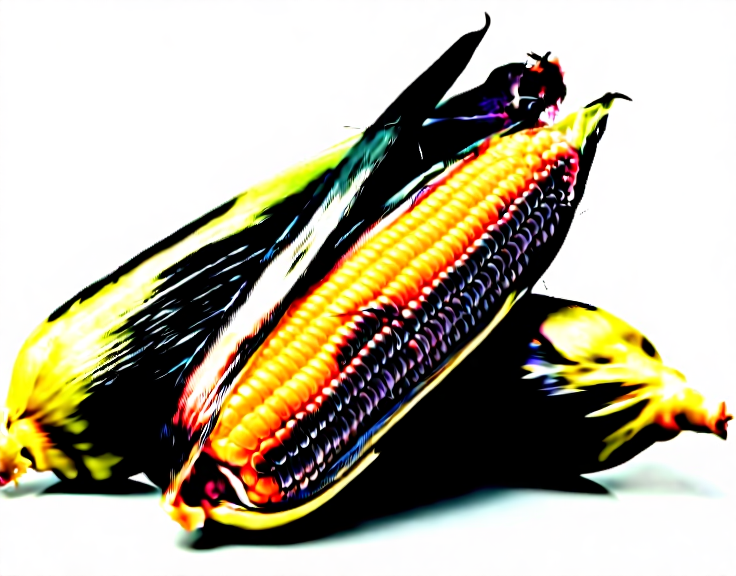} \hspace{-5mm}  &
\includegraphics[width=0.15\textwidth, height=0.08\textheight]{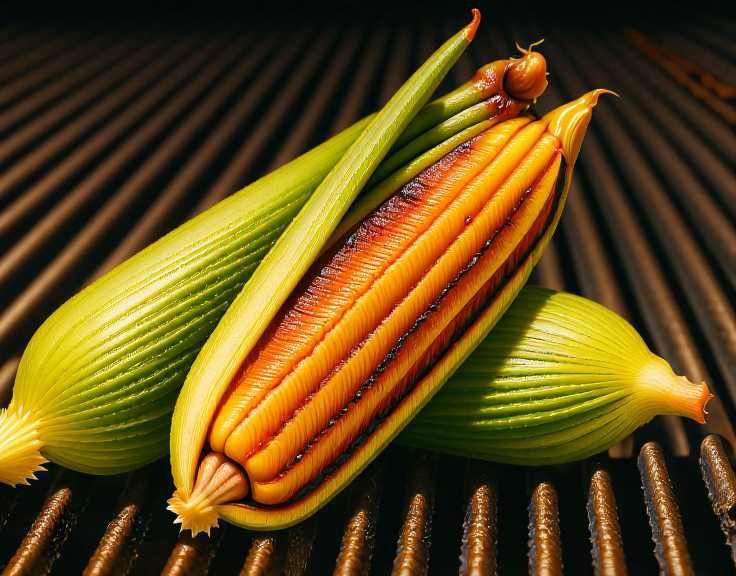} \hspace{-5mm}  &
\includegraphics[width=0.15\textwidth, height=0.08\textheight]{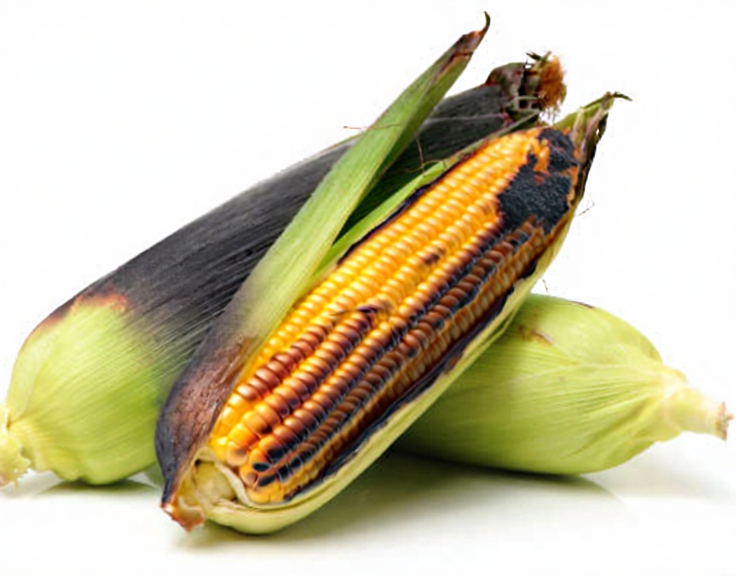} \hspace{-5mm}  
\\
\includegraphics[width=0.15\textwidth, height=0.08\textheight]{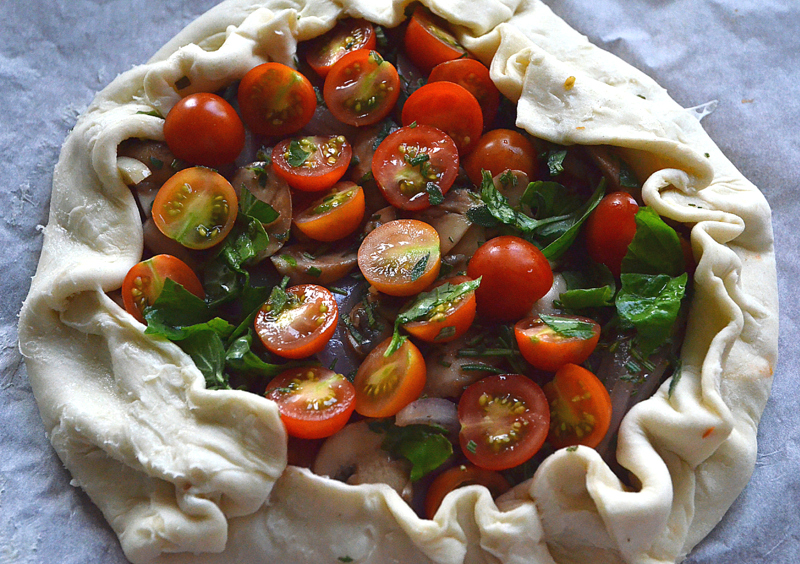} \hspace{-5mm}  &
\includegraphics[width=0.15\textwidth, height=0.08\textheight]{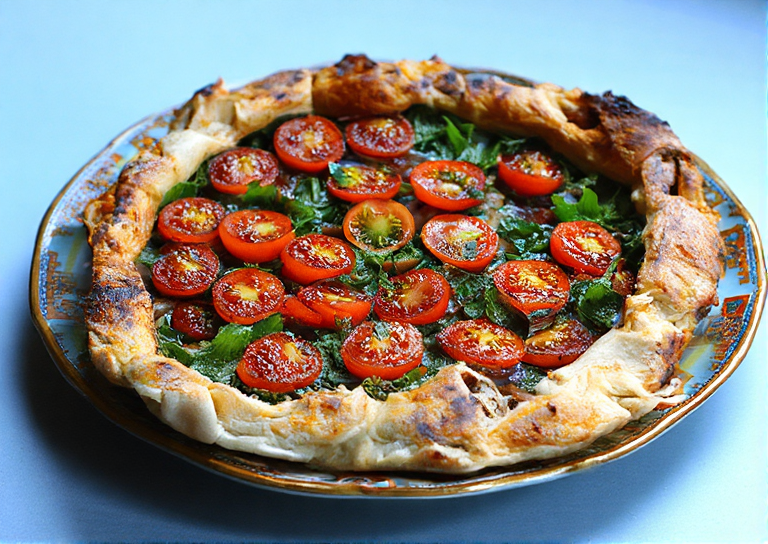} \hspace{-5mm}  &
\includegraphics[width=0.15\textwidth, height=0.08\textheight]{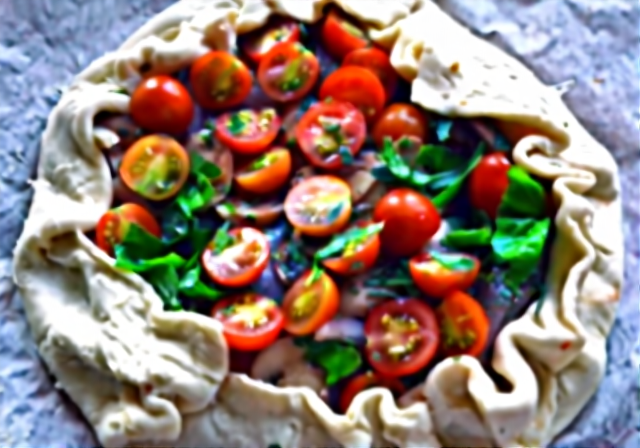} \hspace{-5mm}  &
\includegraphics[width=0.15\textwidth, height=0.08\textheight]{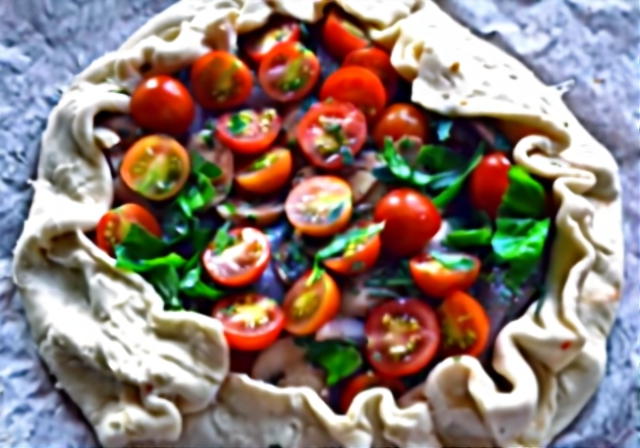} \hspace{-5mm}  &
\includegraphics[width=0.15\textwidth, height=0.08\textheight]{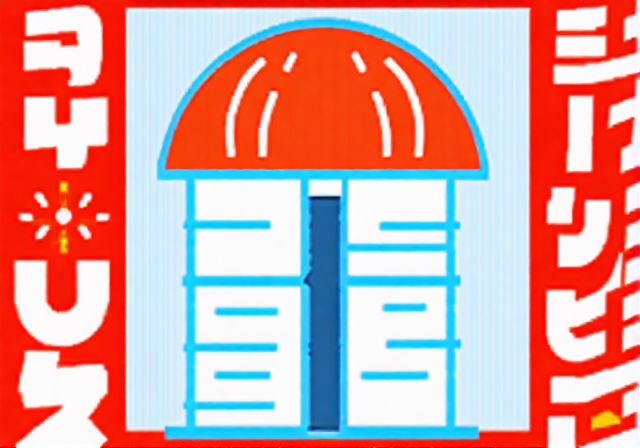} \hspace{-5mm}  &
\includegraphics[width=0.15\textwidth, height=0.08\textheight]{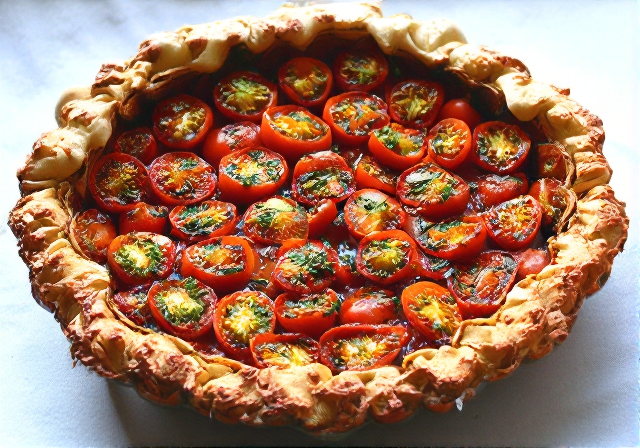} \hspace{-5mm}  
\\
\includegraphics[width=0.15\textwidth, height=0.08\textheight]{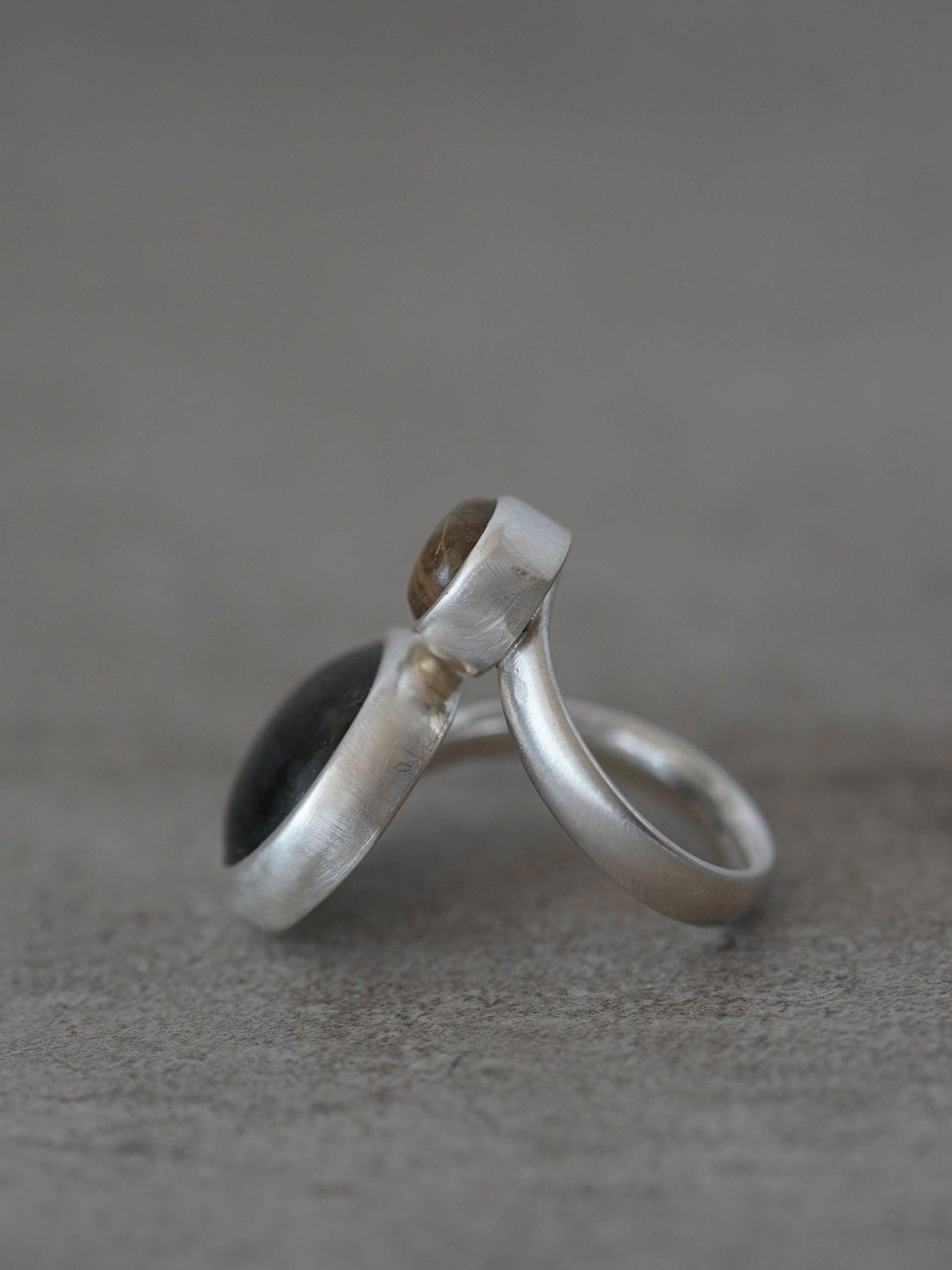} \hspace{-5mm}  &
\includegraphics[width=0.15\textwidth, height=0.08\textheight]{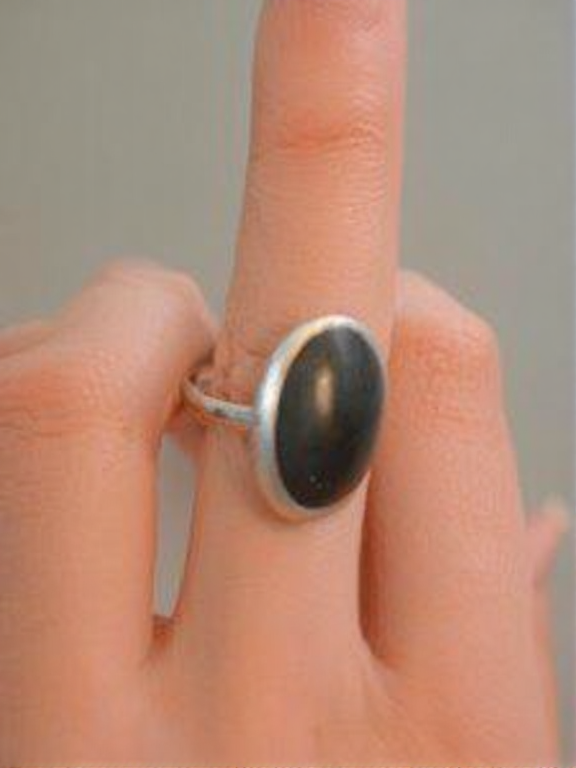} \hspace{-5mm}  &
\includegraphics[width=0.15\textwidth, height=0.08\textheight]{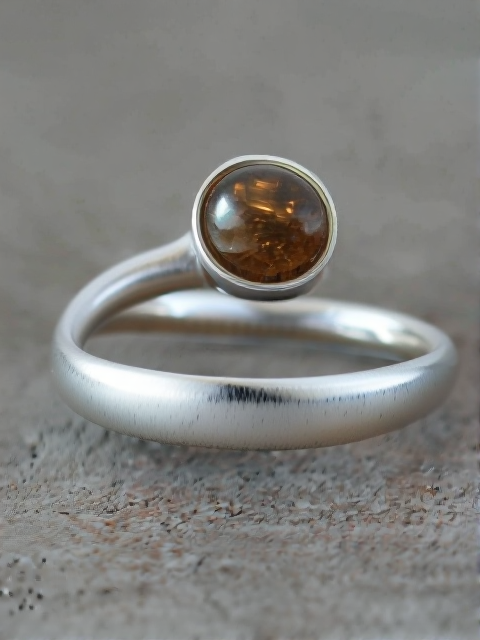} \hspace{-5mm}  &
\includegraphics[width=0.15\textwidth, height=0.08\textheight]{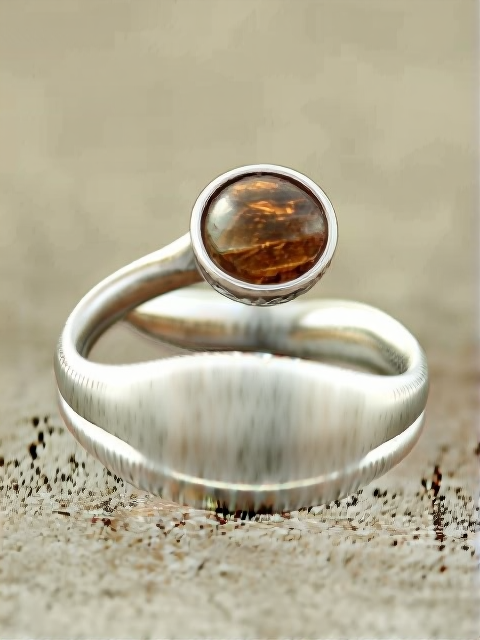} \hspace{-5mm}  &
\includegraphics[width=0.15\textwidth, height=0.08\textheight]{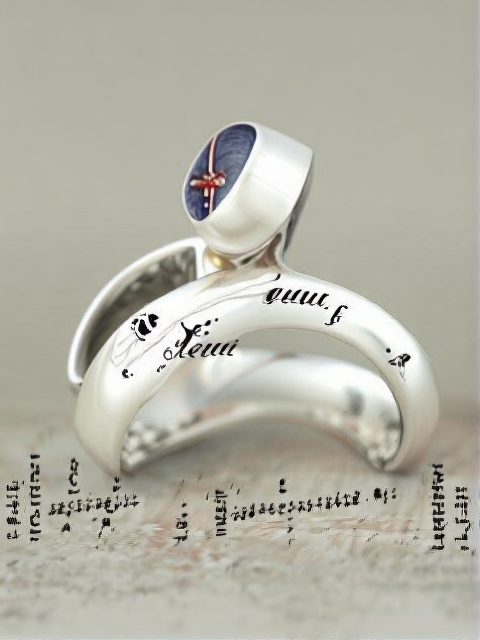} \hspace{-5mm}  &
\includegraphics[width=0.15\textwidth, height=0.08\textheight]{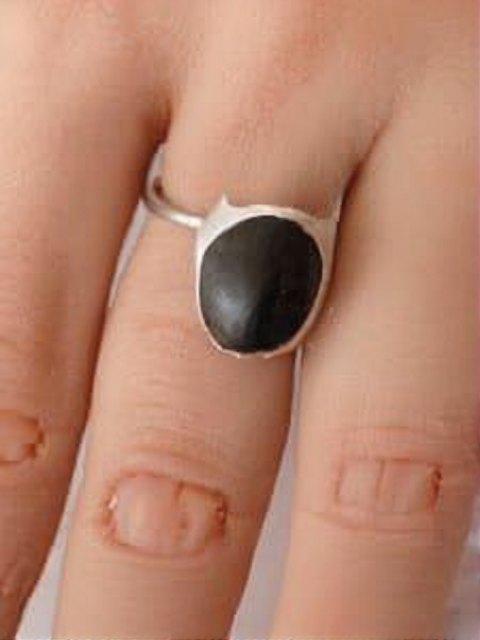} \hspace{-5mm}  
\\
\quad Input & \quad Vanilla & \quad FastV~\cite{chen2024image} & \quad PDrop~\cite{xing2025conical} & \quad IVC-Prune~\cite{sun2026ivc} & \quad \textbf{G$^2$TR (ours)} 

\end{tabular}
}
\vspace{-3mm}
\caption{\small{Visualization of Editing results. The instructions of the four groups of images from top to down are: ``Draw what it will look like after being frozen.''; ``Draw what it will look like after one hour on a hot grill.''; ``What does this dish look like after it has been baked?''; ``Can I see the appearance of this ring on a finger?''. Generally, G$^2$TR performs better in UMMs for image editing than compared baselines.}}
\label{fig:edit_visual}
\vspace{-6mm}
\end{figure}

\vspace{-3mm}
\subsection{Main Performance}
\vspace{-3mm}
\textbf{Performance on image editing tasks.} Figure~\ref{fig:edit_visual} reveals some image editing results of G$^2$TR and baselines. Images and instructions are from RISE and Intell. Compared with the original results, several baselines produce blurred details or fail to follow the editing instructions. In more severe cases, the edited images deviate greatly from the input image, such as the result of FastV in the first group and the result of IVC-prune in the third group. In contrast, after applying G$^2$TR, the model can largely follow the instructions and complete the edits successfully. It also preserves an editing quality comparable to the vanilla model setting. This supports the core claim in Section~\ref{sec:insight}.

\textbf{Performance on image understanding tasks.} Table~\ref{tab:und} reports the results on image understanding benchmarks. G$^2$TR achieves the best relative average among all baselines on both UMMs, reaching 94.9\% on InternVL-U and 99.0\% on BAGEL-7B while keeping only 50\% visual tokens. 
On BAGEL-7B, G$^2$TR matches the vanilla performance on MMBench, obtains the best result on MMVP, and gives the smallest drop on RealWorldQA. On InternVL-U, it also achieves the best or tied-best results on three benchmarks among all compared baselines. 
These results show that our reduction does not sacrifice understanding ability.  
Instead, the VAE latent guidance may help retain several  visual tokens that are useful for both generation and understanding. This may be because the generation and understanding branches in UMMs can provide complementary signals. Generally, performances on image understanding and editing provide positive answers to question (1) in Section~\ref{sec:experiment_settings}.

\begin{table}[t]
    \centering
    \caption{Performance of G$^2$TR compared with baselines on several understanding benchmarks. ``Avg.'' means average and ``Rel. Avg.'' means relative average. We \textbf{highlight} the best performances and \underline{underline} the second best ones. Generally, G$^2$TR reaches the best relative average across both UMMs.}
    \vspace{-2mm}
    \resizebox{\textwidth}{!}{
    \begin{tabular}{clc|cccc|c}
    \toprule
      \textbf{Models} & \textbf{Methods}  & \textbf{Avg. Tokens} & \textbf{MME} & \textbf{MMBench (dev)} & \textbf{MMVP} & \textbf{RWQA} & \textbf{Rel. Avg.} \\
    \midrule
    \rowcolor{cyan!8}  \multirow{7}{*}{\textbf{InternVL-U}}   & \texttt{Vanilla} & 100\% & 2057 & 80.5 & 46.0 & 48.8 & 100\% \\
       & FastV~\cite{chen2024image}\texttt{[ECCV'24]} & 54\% & \textbf{2037} & 77.9 & 38.0 & 45.5 & 92.9\% \\
       & W-FastV~\cite{wen2025token}\texttt{[ACL'25]} & 54\% & 1926 & \underline{78.6} & \underline{39.3} &  45.2 & 92.3\% \\
       & PDrop~\cite{xing2025conical}\texttt{[CVPR'25]} & 61\% & 1953 & 77.7 & 38.7 & 45.6 & 92.0\% \\
       & VSCAN~\cite{zhangvscan}\texttt{[TMLR'26]} & 50\% & 1968 & 78.1 & \textbf{40.0} & 43.5 & 92.2\% \\
       
       & IVC-Prune~\cite{sun2026ivc}\texttt{[ICLR'26]} & 50\% & 2020 & 78.4 & 39.0 & \underline{46.0} & \underline{93.7\%} \\
       & \textbf{G$^2$TR (ours)} & 50\% & \underline{2034} & \textbf{78.8} & \textbf{40.0} & \textbf{46.8} & \textbf{94.9\%} \\
       \midrule
     \rowcolor{cyan!8}   \multirow{7}{*}{\textbf{BAGEL-7B}}  &  \texttt{Vanilla} & 100\% & 2388 & 88.5 & 69.3 & 60.2  & 100\% \\
       & FastV~\cite{chen2024image}\texttt{[ECCV'24]} & 54\% & 2306 & 85.7 & 68.0 & 55.2 & 95.8\% \\
       & W-FastV~\cite{wen2025token}\texttt{[ACL'25]} & 54\% & 2287 & 85.9 & 67.3 & 55.4 & 95.6\% \\
       & PDrop~\cite{xing2025conical}\texttt{[CVPR'25]} & 61\% & 2312 & 84.0 & 66.7 & 53.8 & 94.1\% \\
       & VSCAN~\cite{zhangvscan}\texttt{[TMLR'26]} & 50\% & \textbf{2339} & \textbf{88.5} & \underline{68.7} & 48.4 & 94.4\% \\
       & IVC-Prune~\cite{sun2026ivc}\texttt{[ICLR'26]} & 50\% & \underline{2332} &  \underline{88.4} & \textbf{70.0} & \underline{57.8} & \underline{98.6\%} \\
       & \textbf{G$^2$TR (ours)} & 50\% & 2318 & \textbf{88.5} & \textbf{70.0} & \textbf{59.0} & \textbf{99.0\%} \\
    \bottomrule
    \end{tabular}
    }
    \label{tab:und}
    \vspace{-3mm}
\end{table}

\noindent \textbf{Efficiency.} Table~\ref{tab:efficiency} reports efficiency of our method compared with baselines. G$^2$TR reduces the average visual tokens to 50\%, leading to a 1.90$\times$ KV-cache reduction and a 1.94$\times$ prefill FLOPs speedup over the vanilla model. Compared with FastV, G$^2$TR further lowers the FLOPs ratio from 55.10\% to 51.64\%, while using fewer tokens. It also clearly outperforms IVC-Prune under the same token budget, reducing prefill FLOPs from 3.0619T to 2.1341T. The decode latency is slightly improved to 47.88 ms/token, showing that our method brings consistent efficiency gains without introducing extra decoding overhead, answering question (2) in Section~\ref{sec:experiment_settings}.

\begin{table}[t]
\centering
\caption{Efficiency comparison on BAGEL-7B-MoT. Size of KV Cache, average visual tokens (Avg. Tokens), FLOPs ratio, prefill FLOPs and decode latency are evaluated.}
\vspace{-2mm}
\label{tab:efficiency}
\small
\renewcommand{\arraystretch}{1.15}
\resizebox{\linewidth}{!}{
\begin{tabular}{l|c|ccccc}
\toprule
\multirow{2}{*}{\textbf{Model}} & \textbf{Method} & \textbf{KV Cache}  & \textbf{Avg. Tokens} 
& \textbf{FLOPs Ratio} 
& \textbf{Prefill FLOPs} 
& \textbf{Decode Latency} \\
&  &  \textbf{(MB) $\downarrow$} & \textbf{(\%) $\downarrow$} & \textbf{$\downarrow$} & \textbf{(T) $\downarrow$} & \textbf{(ms/token) $\downarrow$}\\
\midrule
\rowcolor{gray!8} \multirow{4}{*}{BAGEL-7B}
 & Vanilla & 42.02 (1.00$\times$)
& 100\% (1.00$\times$) 
& 100.00\% (1.00$\times$)
& 4.1325 (1.00$\times$) 
& 49.60 (1.00$\times$) \\
& FastV
& \textbf{22.04 (1.91$\times$)}
& 54\% (1.85$\times$)
& 55.10 \% (1.85$\times$)
& 2.2769 (1.82$\times$)
& 48.04 (1.03$\times$) \\
&  IVC-Prune
& 22.06 (1.90$\times$)
&  \textbf{50\% (2.00$\times$)}
&  74.09\% (1.37$\times$) 
& 3.0619 (1.35$\times$)
& 49.30 (1.01$\times$) \\
& \textbf{G$^2$TR (Ours)}
& 22.06 (1.90$\times$)
& \textbf{50\% (2.00$\times$)}
& \textbf{51.64\% (1.94$\times$)} 
& \textbf{2.1341 (1.94$\times$)}
&  \textbf{47.88 (1.04$\times$)} \\
\bottomrule
\end{tabular}
}
\vskip -0.15in
\vspace{-2mm}
\end{table}

\begin{table}[t]
    \centering
    \begin{minipage}[b]{0.48\linewidth}
    \centering
    \setlength{\tabcolsep}{3pt}
    \caption{Quantitative results on image editing. We \textbf{highlight} and \underline{underline} the best and second best performances.}
    \resizebox{\linewidth}{!}{
    \begin{tabular}{ll|cccc|c}
    \toprule
      
      \multirow{2}{*}{\textbf{Models}} & \multirow{2}{*}{\textbf{Methods}}    &  \multicolumn{3}{c}{\textbf{GEdit (EN)}} & \multirow{2}{*}{\textbf{Intell.}} & \textbf{Rel.} \\
     & & \textbf{G\_SC} & \textbf{G\_PQ} & \textbf{G\_O} & & \textbf{Avg.} \\
    \midrule
     \rowcolor{cyan!8}   \multirow{4}{*}{\textbf{BAGEL-7B}}  &  \texttt{Vanilla} & 7.36 & 6.83 & 6.52 & 44.9 & 100\% \\
       & FastV & \underline{7.09} & 6.58 & \underline{6.35} & \underline{43.0} & \underline{96.5\%} \\
      & VSCAN & 7.03 & \underline{6.65} & 6.32 & 42.6 & 96.2\% \\
       & IVC-Prune & 6.82 & 6.52 & 6.27 & 41.1 & 94.0\% \\
       & \textbf{G$^2$TR (ours)} & \textbf{7.18} & \textbf{6.67} & \textbf{6.42} & \textbf{44.0} & \textbf{98.0\%} \\
    \bottomrule
    \end{tabular}
    }
    \label{tab:quan_edit}
    \end{minipage}
    \hfill
    \begin{minipage}[b]{0.48\linewidth}
    \centering
    \setlength{\tabcolsep}{3pt}
    \caption{Results when opening thinking mode. We \textbf{highlight} and \underline{underline} the best and second best performances.}
    \resizebox{\linewidth}{!}{
    \begin{tabular}{ll|cc|cc|c}
    \toprule
      
      \textbf{Models} & \textbf{Methods}    & \textbf{MMVP} & \textbf{RWQA} & \textbf{RISE} & \textbf{Intell.} & \textbf{Rel. Avg.} \\
    \midrule
     \rowcolor{cyan!8}   \multirow{4}{*}{\textbf{BAGEL-7B}}  &  \texttt{Vanilla} & 64.7 & 63.4 & 11.9 & 55.3 & 100\% \\
       & FastV & 63.3 & \underline{62.2} & \underline{9.1} & 51.9 & 91.6\% \\
       & VSCAN & \textbf{66.7} & 60.2 & 8.3 & \underline{53.7} & 91.2\% \\
     \textbf{w/ CoT}  & IVC-Prune & 64.0 & 62.1 & 7.6 & 51.4 & 88.4\% \\
       & \textbf{G$^2$TR (ours)} & \underline{66.0} & \textbf{63.4} & \textbf{9.8} & \textbf{54.3} & \textbf{95.6\%} \\
    \bottomrule
    \end{tabular}
    }
    \label{tab:thinking}
    \end{minipage}
    \vspace{-4mm}
\end{table}

\begin{figure*}[t]
    \centering

    \begin{minipage}[t]{0.32\textwidth}
        \centering
        \includegraphics[width=\linewidth]{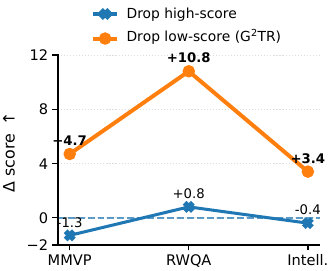}
        \vspace{-2mm}
    \end{minipage}
    \hfill
    \begin{minipage}[t]{0.32\textwidth}
        \centering
        \includegraphics[width=\linewidth]{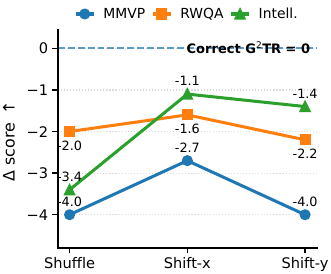}
        \vspace{-2mm}
    \end{minipage}
    \hfill
    \begin{minipage}[t]{0.32\textwidth}
        \centering
        \includegraphics[width=\linewidth]{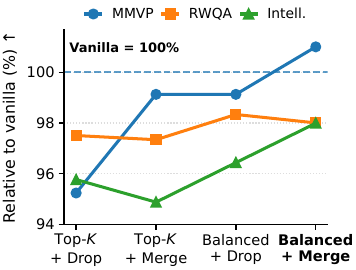}
        \vspace{-2mm}
    \end{minipage}

    \vspace{-2mm}
    \caption{
    Further analysis. (a) Validity of the VAE-alignment score. (b) Effect of perturbing the spatial token-to-anchor relationship. (c) Component ablation of balanced selection and token merging.
    }
    \label{fig:further_analysis}
    \vspace{-5mm}
\end{figure*}

\begin{table}[t]
    \centering
    \renewcommand{\arraystretch}{1.15}
    \caption{Results on different keep ratios. We \textbf{highlight} the best performances under the same keep ratio. G$^2$TR keeps the good performance under ratios of 50\%, 37.5\% and 25\%. }
    \vspace{-2mm}
    \resizebox{\linewidth}{!}{
    \begin{tabular}{l|ccc|ccc|ccc}
    \toprule
      \multirow{2}{*}{\textbf{Benchmarks}}   & \multicolumn{3}{c}{50\%} & \multicolumn{3}{c}{37.5\%} & \multicolumn{3}{c}{25\%}\\
      \cmidrule{2-10}
      & FastV & IVC-Prune & \textbf{G$^2$TR (ours)} & FastV & IVC-Prune & \textbf{G$^2$TR (ours)}  & FastV & IVC-Prune & \textbf{G$^2$TR (ours)}  \\
      \midrule
       \textbf{MMVP}  &  68.0 & \textbf{70.0} & \textbf{70.0} & 60.3 & 58.7 & \textbf{63.3} & 54.7 & 56.0 &  \textbf{60.0} \\
       \textbf{Intell.} & 43.0 & 41.1 & \textbf{44.0} & 41.8 & 41.0 & \textbf{42.5} & 40.3 & 41.0 &  \textbf{42.0} \\
       \bottomrule 
    \end{tabular}
    }
    \vspace{-2mm}
    \label{tab:ratio}
\end{table}

\begin{figure}[t]
    \centering
    \includegraphics[width=\linewidth]{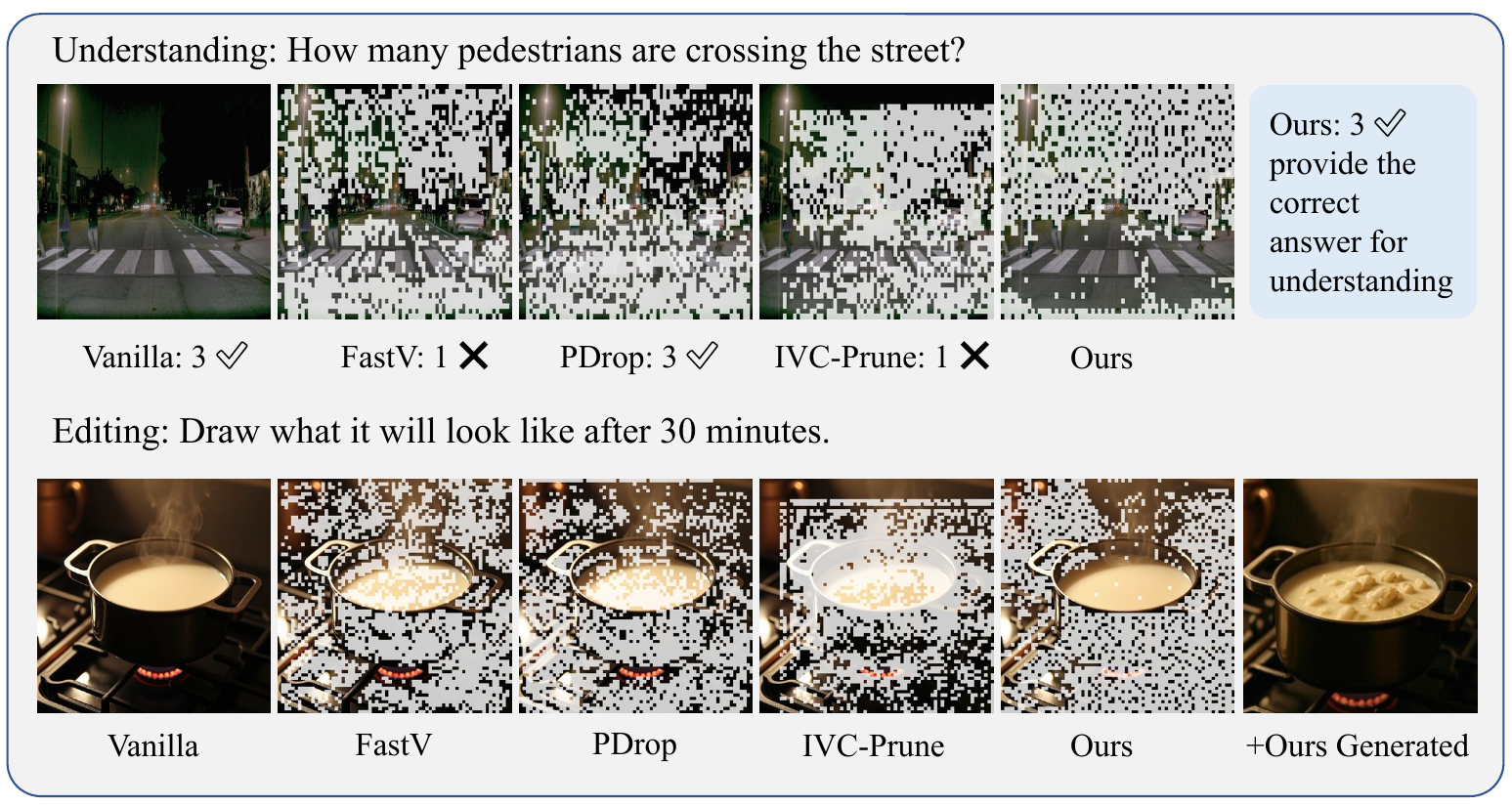}
    \caption{Visualization of retained tokens. The first line illustratively shows one case of image understanding, and the second line shows one case of image editing. }
    \label{fig:token_visual}
    \vspace{-5mm}
\end{figure}

\noindent \textbf{Quantitative results of image editing.} Beyond the direct visual comparisons, we further measure image editing quality with the original evaluation protocols of utilized benchmarks. As shown in Table~\ref{tab:quan_edit}, G$^2$TR also shows clear numerical advantages over existing token reduction baselines. On GEdit-Bench (EN), our method achieves the best results among compressed variants on all three metrics. It also obtains 44.0 on IntelligentBench, close to the vanilla model while using only half of the visual tokens. Overall, G$^2$TR reaches a 98.0\% relative average, outperforming other baselines. 

\noindent \textbf{Thinking mode.} We evaluate the performance when opening the thinking mode of BAGEL-7B-MoT. Following their original report~\cite{deng2025bagel}, we evaluate RISE~\cite{zhaoenvisioning} only when thinking mode is open here. For understanding benchmarks, we set the max response token length to 2048 for avoiding truncation. Table~\ref{tab:thinking} reports the results. We observe no consistent advantage of thinking mode over the non-thinking setting across benchmarks. Under this setting, G$^2$TR still shows a stable trade-off. It reaches the best relative average among all token reduction methods, preserving 95.6\% of the vanilla performance. It also matches the vanilla model on RealworldQA and remains close on IntelligentBench, while keeping competitive results on MMVP. These results indicate that G$^2$TR is still effective when UMMs use thinking mode, and does not rely on a specific decoding style.

\noindent \textbf{Keep ratio.} Table~\ref{tab:ratio} studies the performance under different keep ratios.
G$^2$TR consistently achieves the best or tied-best results across three token budgets on both MMVP and IntelligentBench.
Notably, its advantage remains clear under more aggressive reduction: at a 25\% keep ratio, G$^2$TR reaches 60.0 on MMVP and 42.0 on IntelligentBench, outperforming both FastV and IVC-Prune.
These results demonstrate that G$^2$TR remains robust across different levels of visual token reduction.

\begin{wraptable}{r}{0.45\textwidth}
\centering
\vspace{-6mm}
\caption{Ablation on guidance sources. We \textbf{highlight} the best performances.}
\label{tab:guidance_source_ablation}
\resizebox{\linewidth}{!}{
\begin{tabular}{lccc}
\toprule
\textbf{Guidance Source} & \textbf{MMVP} $\uparrow$ & \textbf{RWQA} $\uparrow$ & \textbf{Intell.} $\uparrow$  \\
\midrule
Random         & 65.3 & 48.2 & 40.6 \\
Attention-based & 68.7 & 55.8 & 42.6  \\
Text similarity & 68.0 & 55.2 & 41.7 \\
\rowcolor{orange!8} \textbf{VAE latent (ours)} & \textbf{70.0} & \textbf{59.0} & \textbf{44.0} \\
\bottomrule
\end{tabular}
}
\vspace{-2mm}
\end{wraptable}

\vspace{-2mm}
\subsection{Ablation study.} 
\vspace{-2mm}
To study whether our guidance source is the better choice compared with others, we make this study. In detail, we also employ three selection signals: random, attention-based importance and image-text feature similarity (text similarity) right after the ViT encoding process. Table~\ref{tab:guidance_source_ablation} reports the results. It can be discovered that using VAE latent as guidance achieves best scores across both understanding and editing tasks. This ablation supports our claim that our proposed guidance plays a core role in the good performance above, compared with other selection signals.

\subsection{Further Analysis}
\label{sec:further_analysis}
We further analyze G$^2$TR from three aspects: score validity, spatial correspondence, and component design. Results are in Figure~\ref{fig:further_analysis}.
First, removing low-score tokens consistently outperforms random removal by 4.7, 10.8, and 3.4 points on MMVP, RWQA, and IntelligentBench, respectively, whereas removing high-score tokens provides no consistent gain over random removal.
This supports the importance ranking induced by VAE alignment.
Second, perturbing the token-to-anchor correspondence through shuffling or spatial shifts consistently degrades performance relative to the original assignment, showing that the gains cannot be explained by arbitrary matching with VAE anchors.
Finally, balanced selection consistently improves direct dropping over global top-$K$ selection.
Token merging has benchmark-dependent effects when used alone, but together with balanced selection it yields the strongest overall performance with improving performances of two benchmarks.
These results further validate the spatial correspondence and individual reduction components of G$^2$TR.

\vspace{-2mm}
\subsection{Visualization of retained tokens.} 
\vspace{-2mm}
We further do this to show what tokens are kept after reduction. As shown in Figure~\ref{fig:token_visual}, the retained tokens preserve visual information useful for both understanding and editing. In the counting example, the retained tokens still cover the pedestrians and allow the model to predict the correct answer. In the editing example, our method preserves most visual tokens of the pot and also visual tokens for the whole layout. The details are important for producing a well-edited image. This suggests that G$^2$TR can keep tokens not only effective for textual understanding, but also friendly to image editing. 

\vspace{-3mm}
\section{Limitations and Future Work}
\label{section:limitation}
\vspace{-3mm}

This work focuses on visual token reduction for separate-encoder UMMs, where the generation branch can provide a natural guidance signal. As UMM architectures continue to evolve, future models may adopt more native and adaptive interactions between understanding and generation pathways. Extending generation-guided reduction to these broader designs is an interesting future direction, which is not explored in our paper. In addition, this paper studies a fixed token budget for clear comparison. A more dynamic strategy that adjusts the reduction ratio based on task difficulty, image content, or latency constraints may further improve the trade-off between efficiency and performance. We believe such directions will help make UMMs more practical for interactive and resource-constrained applications.

\section{Conclusion}
\vspace{-3mm}

In this work, 
we propose G$^2$TR, a training-free and plug-and-play framework that uses the generation branch as guidance for reducing understanding-side visual tokens for UMMs. Different from previous works mainly using text-centric guidance, G$^2$TR preserves visual tokens that are more consistent with generation-related latent representations. Our design reduces the token budget before inner LLM prefill while keeping the attention operators unchanged.
Extensive experiments show that G$^2$TR improves inference efficiency and maintains strong performance on both understanding and image editing tasks. These results suggest that G$^2$TR is a simple and effective direction for efficient UMM inference, thereby benefiting UMM deployment and utilization under resource-constrained scenarios. We hope that G$^2$TR can facilitate the new research trends on UMM efficiency.

\section*{Acknowledgements}
\vspace{-3mm}
This work was supported by Huawei Explore X funds. This work was also sponsored by Al Hundred Schools Program and was carried out using the Ascend AI technology stack.

\noindent \textbf{Competing interests.} The authors declare no additional competing interests beyond the affiliations and funding disclosed above.

\bibliographystyle{plain}
\bibliography{custom}

\clearpage


\appendix

\section{Relationship among Generation, Editing and Understanding-side Visual Tokens}
\label{appendix:relationship}
Image generation (here ``generation'' is a task name) and image editing are two sub-tasks in UMMs. Image generation creates an image from an instruction without a given image. Image editing takes a source image and an instruction, and modifies the source image according to the instruction. Both tasks finally rely on the generation branch (here ``generation'' means a branch name) to produce image results. The key difference lies in the input path. Pure image generation does not involve understanding-side visual tokens, since no source image is encoded by the vision encoder. In contrast, image editing also represents the source image as understanding-side visual tokens, which are then used to condition the generation process. 

Therefore, among these two tasks, only image editing is directly affected by understanding-side visual token reduction. For simplicity, when we mention ``visual tokens'', we are talking about understanding-side visual tokens.

\section{Notations in this paper}
\label{appendix:notations}
We list all the notations in our main paper in Table~\ref{tab:notations}.

\begin{table}[h]
    \centering
    \caption{Notations used in our paper}
    \resizebox{\linewidth}{!}{
    \begin{tabular}{cc}
    \toprule
       Notations  & Meanings \\
    \midrule
       $x$  & input image  \\
       $E_\mathrm{u}$ & understanding encoder (usually ViT) \\
       $\mathbf U$ & the understanding-side visual token sequence \\
       $N_u$ & the number of understanding-side visual tokens \\
       $ \mathbf u_i$ & each visual token \\
       $ \mathbf H_\mathrm{u}, \ \mathbf H_\mathrm{g}$ & hidden states from understanding branch and generation branch \\
       $ P_\mathrm{u}, \ P_\mathrm{g}$ & MLP projectors of the two branches \\
       $ \mathbf P_\mathrm{u}^{\mathrm{pos}} $ & understanding visual position embeddings \\
       $E_\mathrm{g}$ & generation encoder (usually VAE) \\
       $\mathbf Z$ & the generation-side latent token sequence \\
       $ N_g $ & the number of generation-side latent tokens \\
       $ \mathbf z_j$ & each latent token \\
       $ \mathbf P_\mathrm{g}^{\mathrm{pos}} $ & generation-side position embeddings \\
       $ \mathbf e(t) $ & time embeddings \\
       $ \mathbf H_\mathrm{text} $ & hidden states from text instructions \\
       $ (p_i, q_i) $ & the position of a visual token \\
       $ c(i) $ & the position of a latent anchor corresponding to a visual token \\
       $ H_u, \ W_u, \ H_g, \ W_g $ & the height and width of grids \\
       $ \lfloor \rfloor $ & floor option \\
       $ \mathbf{z}_{c(i)} $ & matched VAE latent \\
       $\rho$ & keep ratio \\
       $K$ & token budget \\
       $s_i$ & importance score \\
       $b_\mathbf{c}$ & the best token inside each non-empty latent anchor \\
       $\mathcal{B}$ & the candidate set \\
       $\mathcal{S}, \ \mathcal{S}_0$ & retained tokens \\
       $\bar{\mathcal{S}}$ & redundant tokens \\
       $n(i)$ & index of the retained token nearest to removed token $ \mathbf u_i$ \\
       $\lambda$ & merging weight \\
       $\tilde{\mathbf{u}}_j$ & final retained tokens \\
       $\tilde{\mathbf{U}} $ & final compressed visual token sequence \\
    \bottomrule
    \end{tabular}
    }
    \label{tab:notations}
\end{table}

\section{Detailed Experiment Settings}
\label{appendix:detailed_experiment}

\subsection{Model Usage}

\textbf{BAGEL.} BAGEL~\cite{deng2025bagel} is an open-source UMM that supports both multimodal understanding and visual generation. It uses a decoder-only design and is pretrained on large-scale interleaved text, image, video, and web data. The released BAGEL-7B-MoT model has 7B active parameters and 14B total parameters. Its architecture follows a Mixture-of-Transformer-Experts design, with separate experts and visual encoders for understanding and generation, while sharing self-attention over the same token sequence. This design allows BAGEL to handle tasks such as visual question answering, text-to-image generation, image editing, and free-form visual manipulation in a single model.

\textbf{InternVL-U.} InternVL-U~\cite{tian2026internvl} is a lightweight 4B-parameter UMM for understanding, reasoning, image generation, and image editing. Instead of training a fully native UMM from scratch, it builds on a strong MLLM backbone and adds a specialized MMDiT-based visual generation head. The model follows a unified contextual modeling design with modality-specific modules and decoupled visual encoders. This design aims to preserve the semantic strength of the MLLM while enabling high-quality generation and editing. InternVL-U also uses reasoning data synthesis, including text-rich and scientific tasks, to better align user intent with fine-grained visual outputs.

\subsection{Dataset and Benchmark Usage}

We list the name, the number of samples, and the question types of utilized benchmarks in our main text in Table~\ref{tab:db}.

\begin{table}[h]
    \centering
    \caption{Benchmarks utilized in our main text.}
    \begin{tabular}{ccc}

    \toprule
       Name  & Num Samples & Type \\
    \midrule
       MME~\cite{fumme}  & 2,374 & visual question answering \\
       MMBench (dev)~\cite{liu2024mmbench} & 4,329 & visual question answering \\
       MMVP~\cite{tong2024eyes} & 300 & pair-level visual question answering \\
       RealWorldQA~\cite{xai_grok1_5} & 765 & real-world visual question answering \\
       GEdit (EN)~\cite{liu2025step1x} & 606 & image editing \\
       RISE-Bench~\cite{zhaoenvisioning} &  360   &     reasoning, logical or spatial image editing          \\
       IntelligentBench~\cite{deng2025bagel} & 350 &   reasoning, logical or spatial image editing     \\
    \bottomrule
    \end{tabular}
    
    \label{tab:db}
\end{table}



    

\subsection{Hyper-parameter Choosing}

For all models and baselines, we use the default settings and hyper-parameters described in their original repositories and papers. Notably, we keep the random seed unchanged (42) for all baselines. For G$^2$TR, we keep $\rho$ to 0.5 as described in our paper. We then set $K_{\mathrm{min}}$ to 1. 

For understanding tasks of BAGEL-7B without thinking and InternVL-U, we set the max response token length to 20. For BAGEL with thinking mode, we set the max response token length to 2048 to avoid truncation. For editing tasks, we set the time steps to 50, and other settings are default ones.

\section{Theoretical Analysis}
\label{appendix:theory}
Let $\mathbf U=\{\mathbf u_i\}_{i=1}^{N}$ denote the visual tokens from the understanding encoder, and let
$\mathbf Z=E_{\mathrm{vae}}(\mathbf x)$ denote the latent grid produced by the generation-side VAE.
In unified multimodal models, visual tokens need to support both semantic understanding and visual generation.
 
However, these two objectives may prefer different visual granularity. This motivates recent UMM designs that decouple visual pathways for understanding and generation~\cite{wu2025janus,tian2026internvl}. In this view, the VAE latent space provides a compact representation that is directly optimized for image reconstruction and generation.

We therefore use the generation-side latent as a guidance signal for understanding-side token reduction.
For each ViT token $\mathbf u_i$, we find its corresponding VAE latent anchor $\mathbf z_{c(i)}$, where
$c(i)$ denotes the token-to-latent-anchor mapping. We define the latent-alignment score as
\begin{equation}
    s_i
    =
    \cos\!\left(
        \mathbf u_i,\,
        \mathbf z_{c(i)}
    \right).
\end{equation}
This score is a simple and training-free surrogate for the shared visual factors between the understanding token and the generation latent. Tokens with larger $s_i$ are more consistent with the generation-side representation, and are therefore more likely to preserve visual details that matter for image editing and generation.

Ideally, the retained subset $S$ should preserve the generation-relevant information in $\mathbf U$:
\begin{equation}
    S^\star
    =
    \arg\max_{|S|=K}
    I(\mathbf U_S;\mathbf Z).
\end{equation}
Directly optimizing this objective is intractable during inference. We instead optimize a modular surrogate:
\begin{equation}
    R(S)
    =
    \sum_{i\in S} s_i,
    \qquad |S|=K.
\end{equation}
In practice, we further encourage generation-related information over the VAE latent grid, rather than selecting all tokens from a few highly scored regions. This prevents the retained tokens from collapsing to local salient areas and helps preserve the global image layout.

After token selection, the removed tokens are not simply discarded. Each removed token is assigned to its nearest retained token in feature space and merged into it. Let $\tilde{\mathbf U}_S$ be the compressed token set, and let
$\mathcal R_S(\tilde{\mathbf U}_S)$ denote its expanded approximation in the original token space. The induced token reconstruction error is
\begin{equation}
    \epsilon_{\mathrm{merge}}
    =
    \left\|
        \mathbf U -
        \mathcal R_S(\tilde{\mathbf U}_S)
    \right\|_2 .
\end{equation}
This term measures how well the compressed tokens preserve the original visual token set after merging.

We can then state a conditional perturbation bound. Suppose the following UMM decoder $F(\cdot)$ is locally $L$-Lipschitz with respect to its visual-token input. Then
\begin{equation}
    d\!\left(
        F(\mathbf U),
        F(\mathcal R_S(\tilde{\mathbf U}_S))
    \right)
    \le
    L
    \left\|
        \mathbf U -
        \mathcal R_S(\tilde{\mathbf U}_S)
    \right\|_2 ,
\end{equation}
where $d(\cdot,\cdot)$ denotes the output discrepancy. This bound shows that the output change is controlled by the token reconstruction error. Moreover, the selection score $R(S)$ controls which information is preserved before merging: VAE-guided selection keeps tokens that are more aligned with the generation latent, while token merging reduces the feature distortion caused by removing the remaining tokens.

Therefore, the method provides a generation-aware form of token reduction. Unlike attention-only or text-similarity-based pruning, which mainly preserves tokens useful for language-side prediction, VAE-guided reduction uses the generation branch itself as the relevance signal. This makes the retained tokens more consistent with the visual factors required by image generation, while still reducing the number of understanding-side visual tokens before LLM prefill.

\section{More Visualization Results}
\label{appendix:more_visualization}

\begin{figure}[t]
\scriptsize
\centering
\resizebox{\linewidth}{!}{

\begin{tabular}{cccccc}
\includegraphics[width=0.15\textwidth, height=0.08\textheight]{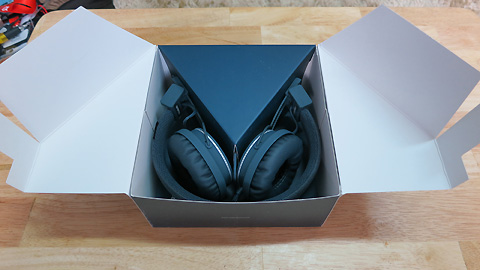} \hspace{-5mm}  &
\includegraphics[width=0.15\textwidth, height=0.08\textheight]{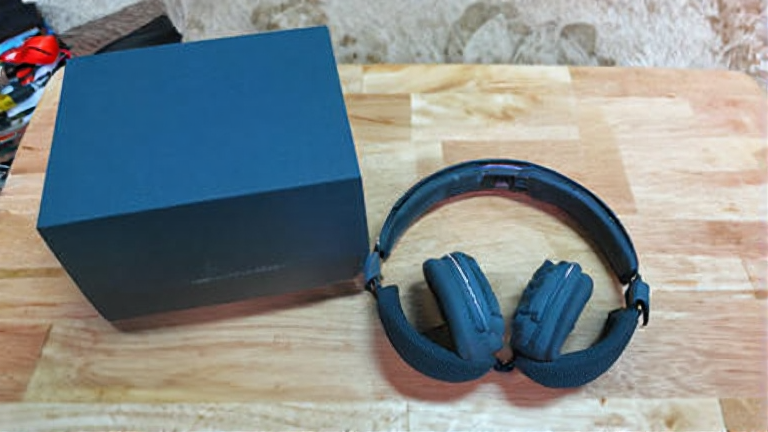} \hspace{-5mm}  &
\includegraphics[width=0.15\textwidth, height=0.08\textheight]{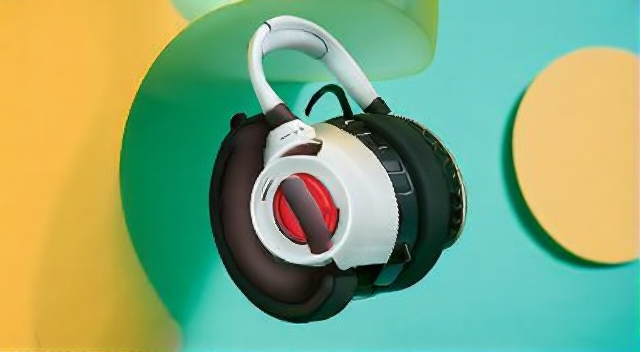} \hspace{-5mm}  &
\includegraphics[width=0.15\textwidth, height=0.08\textheight]{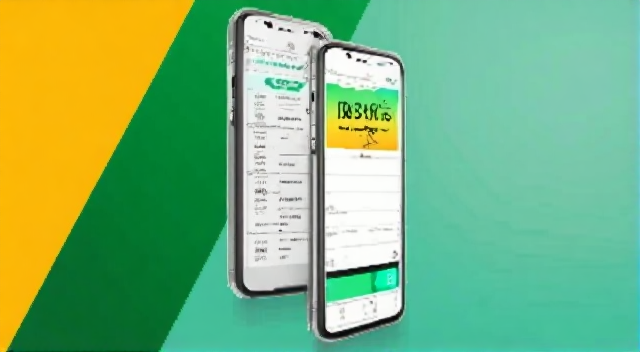} \hspace{-5mm}  &
\includegraphics[width=0.15\textwidth, height=0.08\textheight]{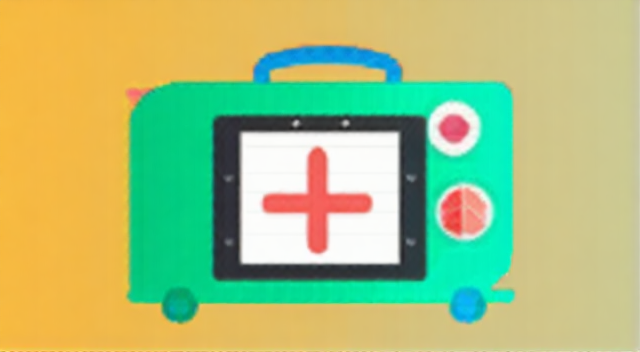} \hspace{-5mm}  &
\includegraphics[width=0.15\textwidth, height=0.08\textheight]{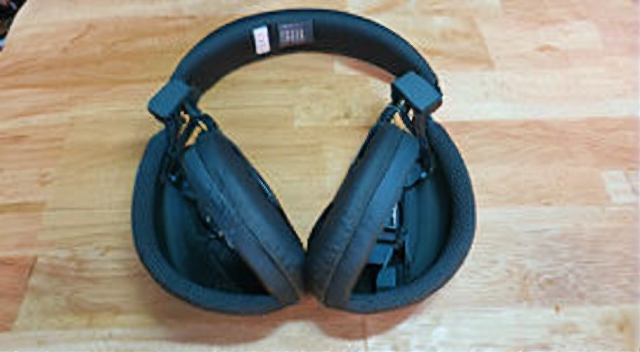} \hspace{-5mm}  
\\
\includegraphics[width=0.15\textwidth, height=0.08\textheight]{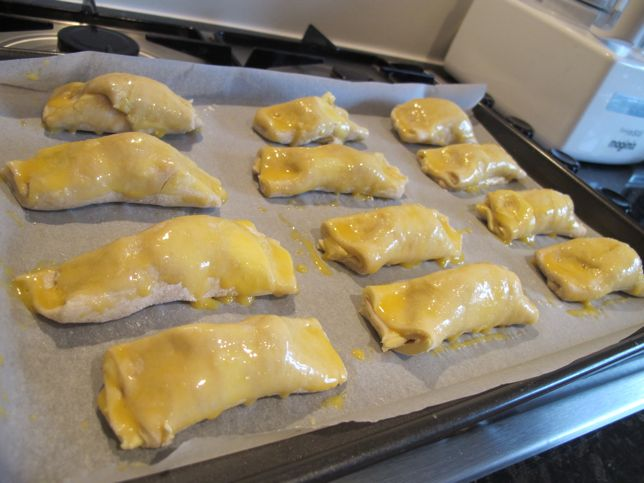} \hspace{-5mm}  &
\includegraphics[width=0.15\textwidth, height=0.08\textheight]{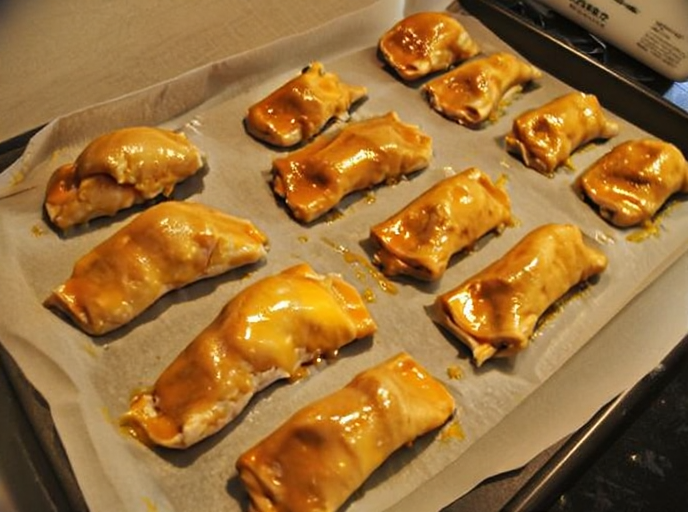} \hspace{-5mm}  &
\includegraphics[width=0.15\textwidth, height=0.08\textheight]{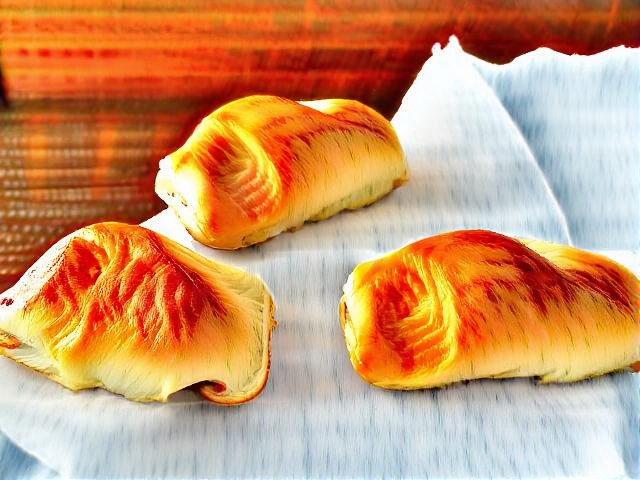} \hspace{-5mm}  &
\includegraphics[width=0.15\textwidth, height=0.08\textheight]{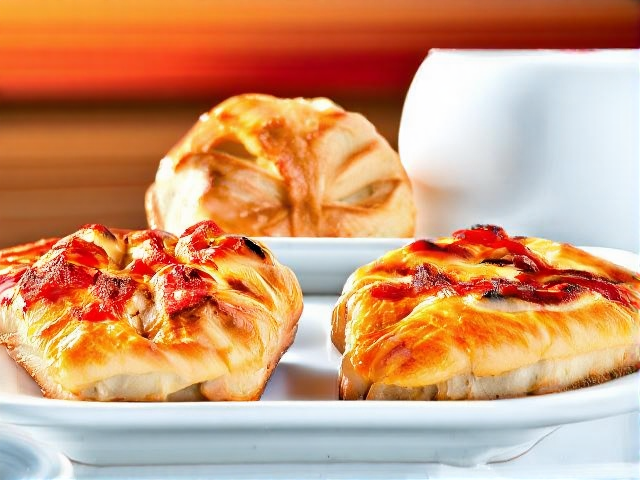} \hspace{-5mm}  &
\includegraphics[width=0.15\textwidth, height=0.08\textheight]{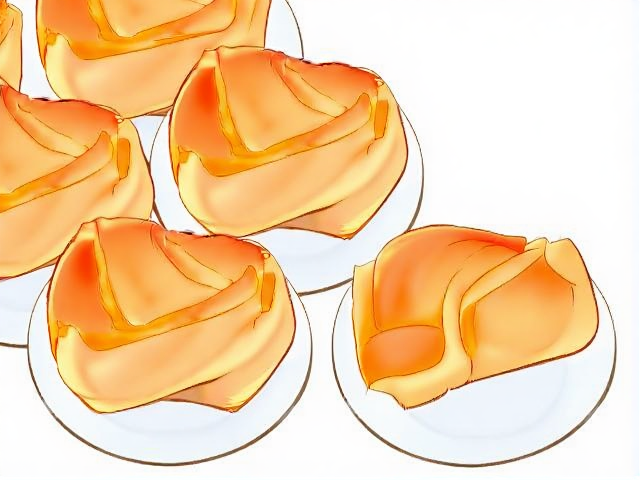} \hspace{-5mm}  &
\includegraphics[width=0.15\textwidth, height=0.08\textheight]{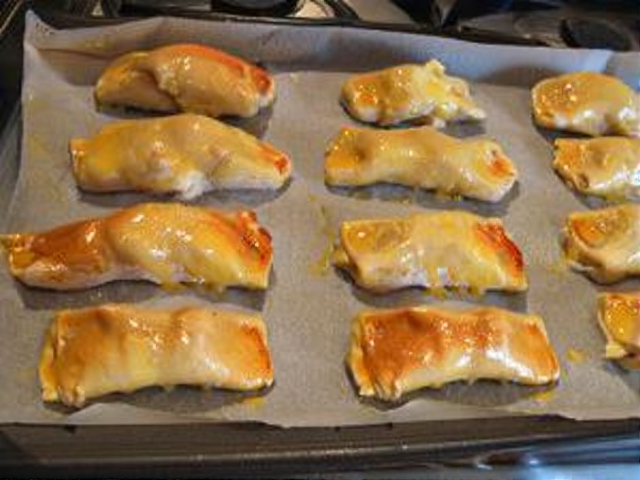} \hspace{-5mm}  
\\
\includegraphics[width=0.15\textwidth, height=0.08\textheight]{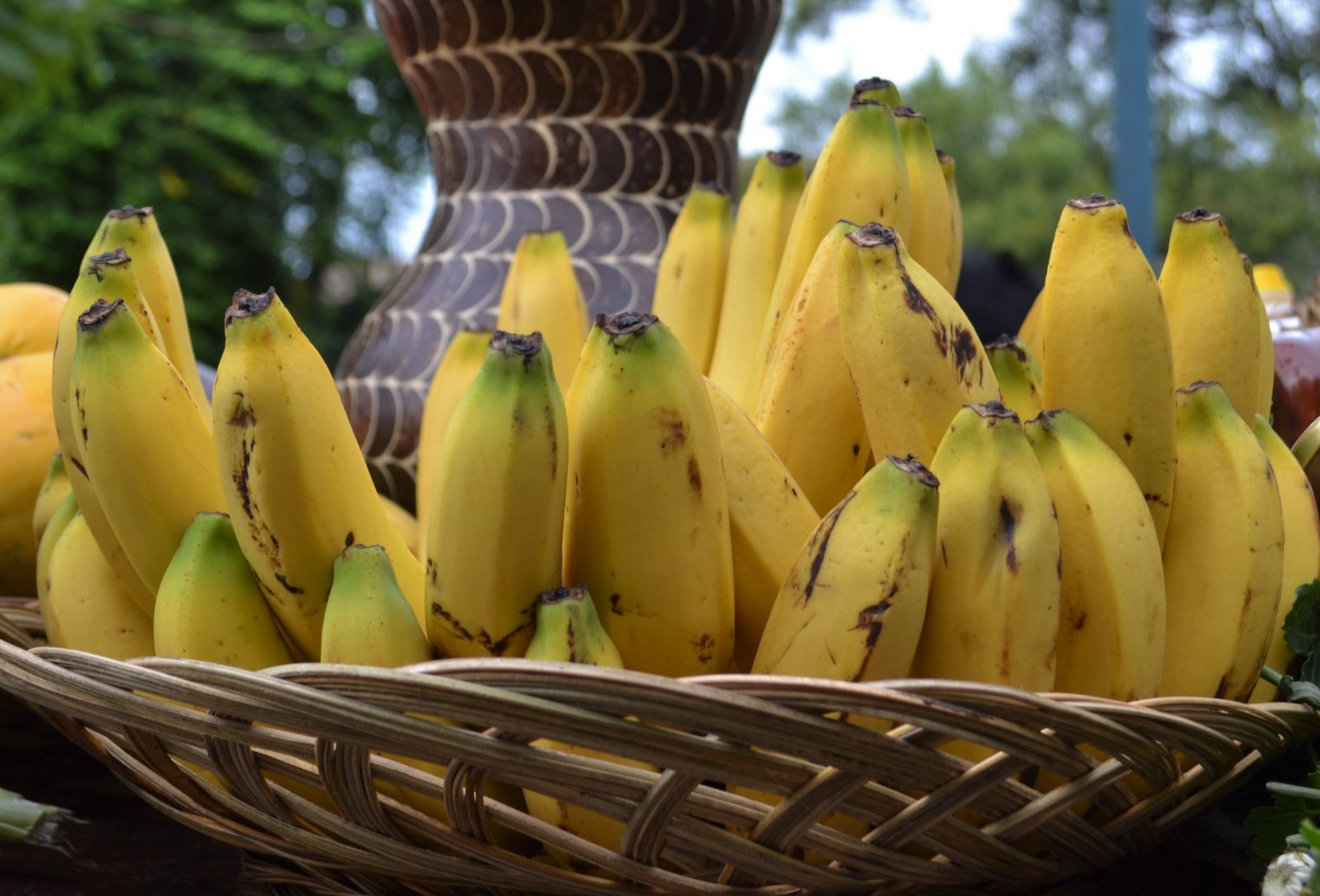} \hspace{-5mm}  &
\includegraphics[width=0.15\textwidth, height=0.08\textheight]{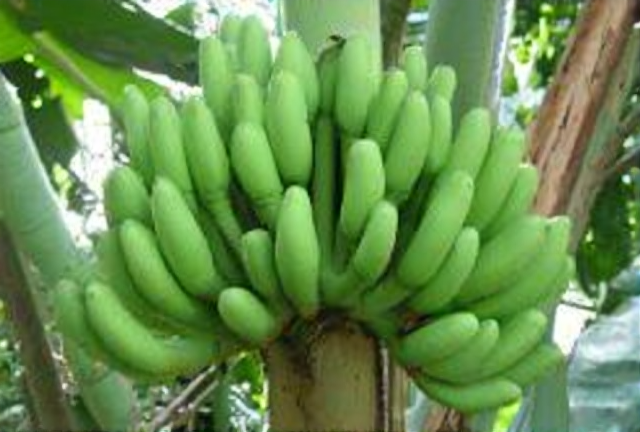} \hspace{-5mm}  &
\includegraphics[width=0.15\textwidth, height=0.08\textheight]{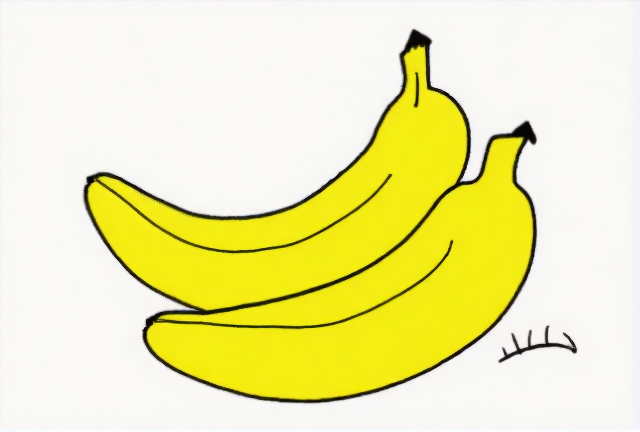} \hspace{-5mm}  &
\includegraphics[width=0.15\textwidth, height=0.08\textheight]{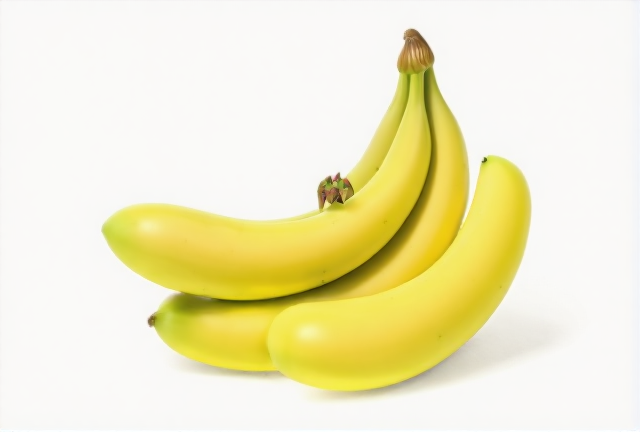} \hspace{-5mm}  &
\includegraphics[width=0.15\textwidth, height=0.08\textheight]{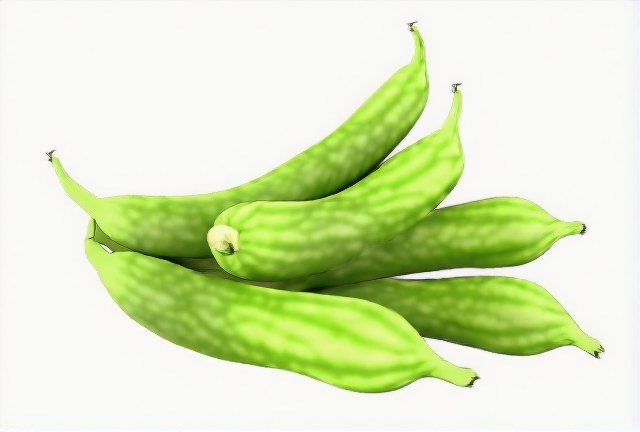} \hspace{-5mm}  &
\includegraphics[width=0.15\textwidth, height=0.08\textheight]{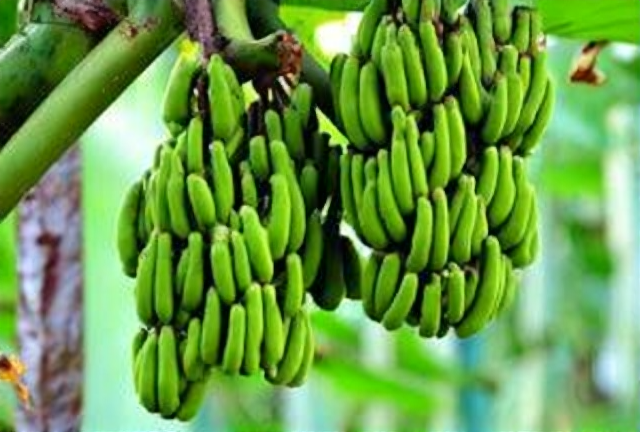} \hspace{-5mm}  
\\
\includegraphics[width=0.15\textwidth, height=0.08\textheight]{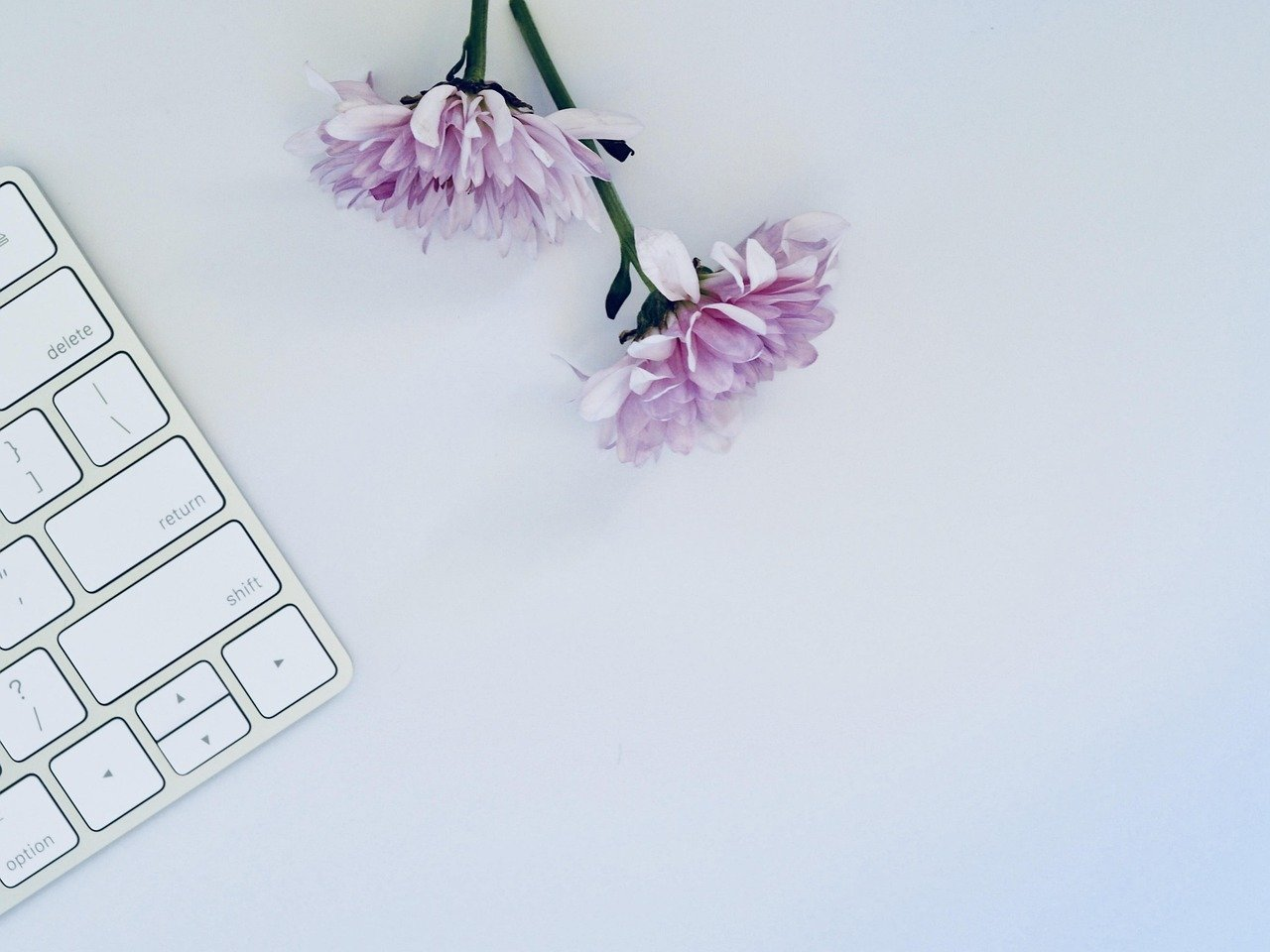} \hspace{-5mm}  &
\includegraphics[width=0.15\textwidth, height=0.08\textheight]{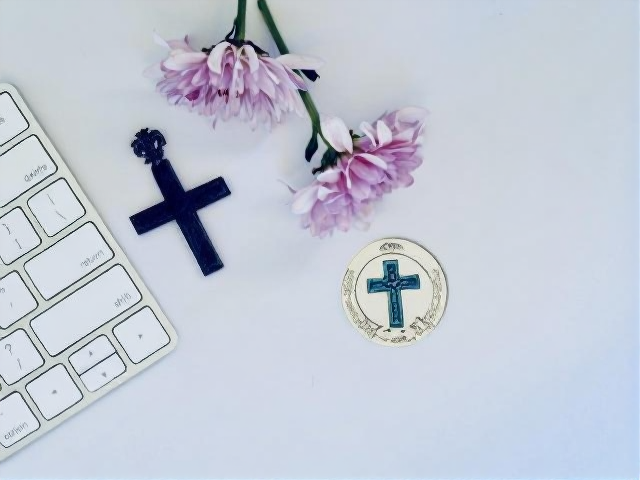} \hspace{-5mm}  &
\includegraphics[width=0.15\textwidth, height=0.08\textheight]{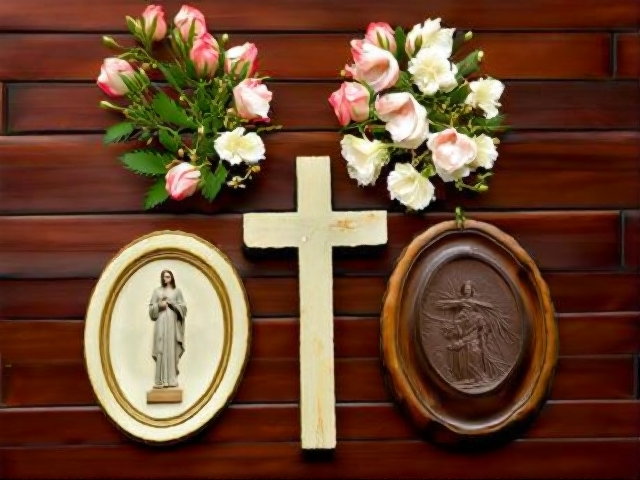} \hspace{-5mm}  &
\includegraphics[width=0.15\textwidth, height=0.08\textheight]{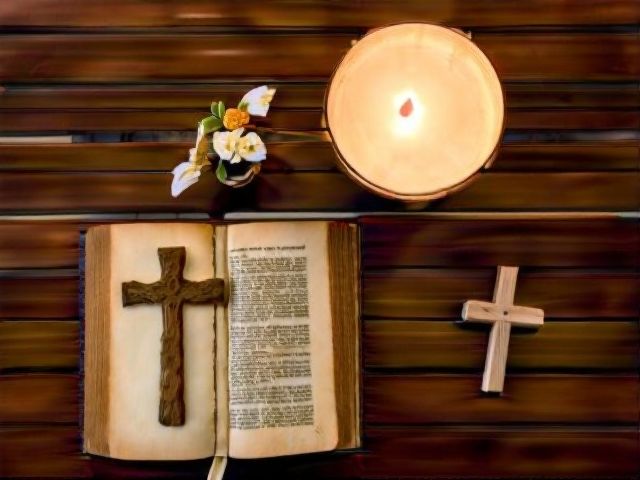} \hspace{-5mm}  &
\includegraphics[width=0.15\textwidth, height=0.08\textheight]{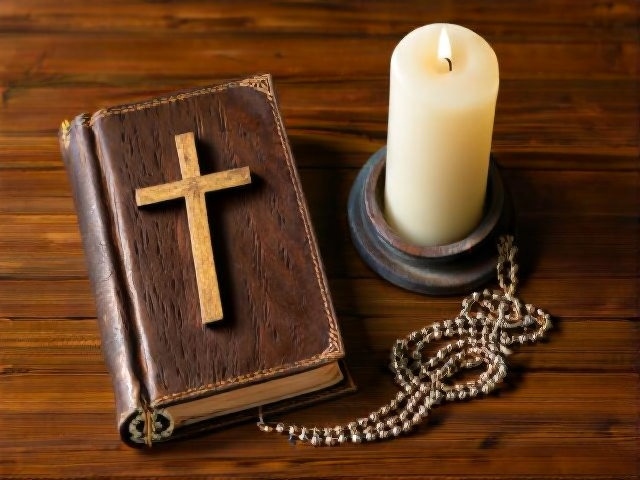} \hspace{-5mm}  &
\includegraphics[width=0.15\textwidth, height=0.08\textheight]{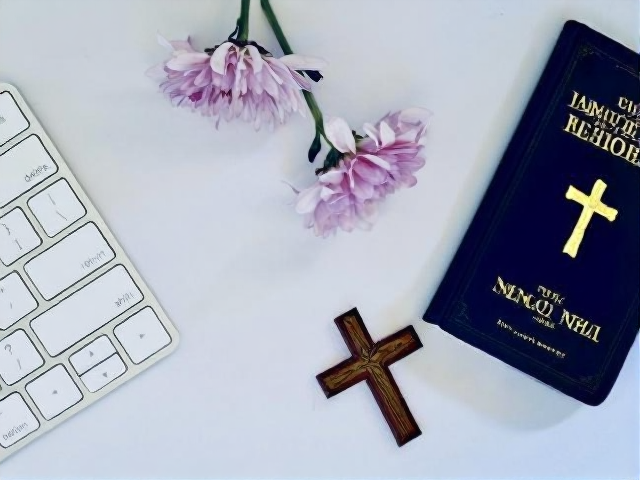} \hspace{-5mm}  
\\
\includegraphics[width=0.15\textwidth, height=0.08\textheight]{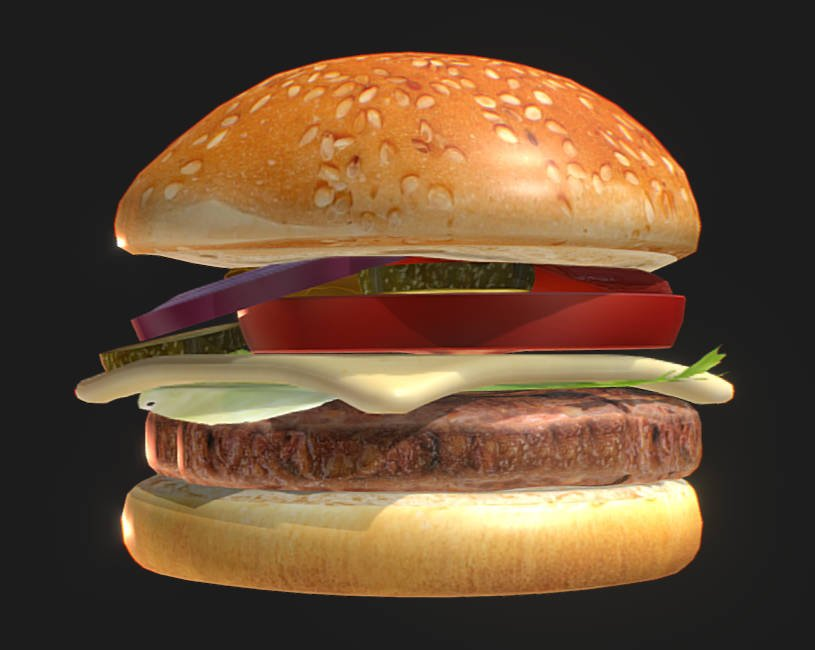} \hspace{-5mm}  &
\includegraphics[width=0.15\textwidth, height=0.08\textheight]{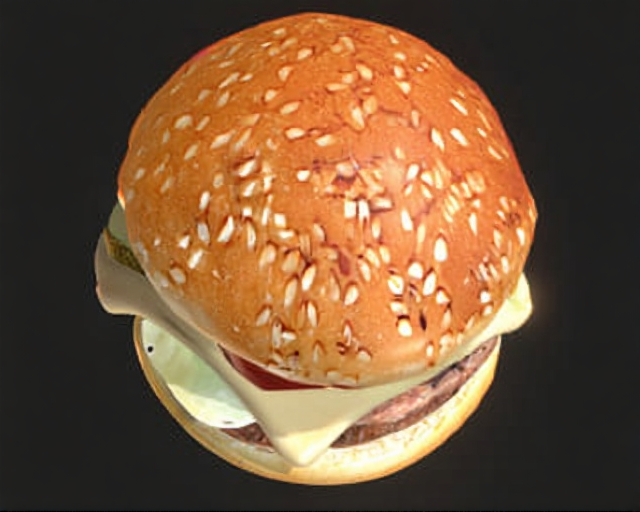} \hspace{-5mm}  &
\includegraphics[width=0.15\textwidth, height=0.08\textheight]{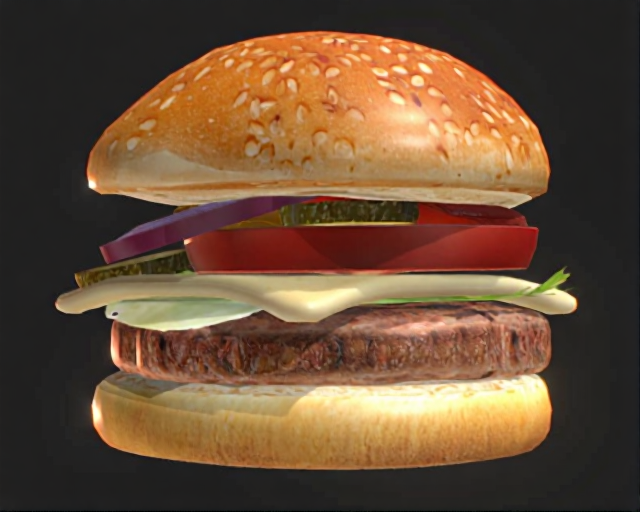} \hspace{-5mm}  &
\includegraphics[width=0.15\textwidth, height=0.08\textheight]{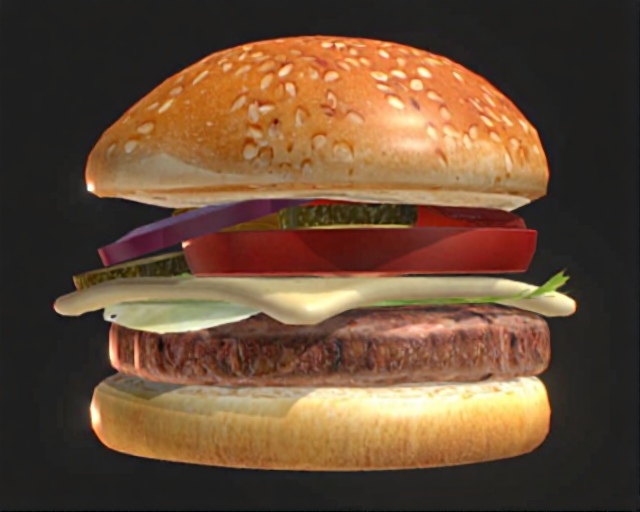} \hspace{-5mm}  &
\includegraphics[width=0.15\textwidth, height=0.08\textheight]{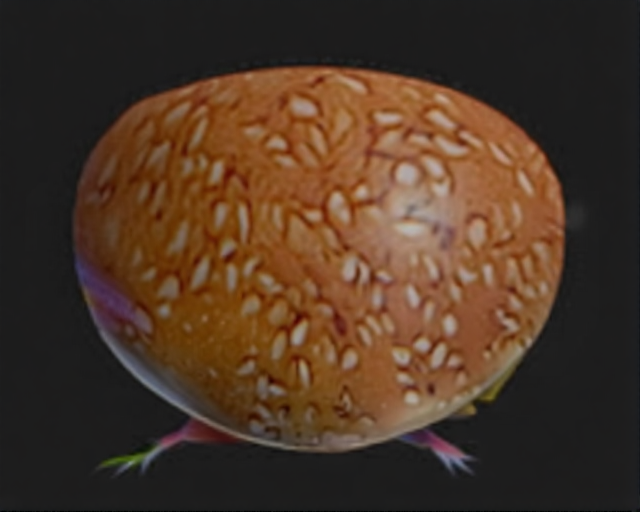} \hspace{-5mm}  &
\includegraphics[width=0.15\textwidth, height=0.08\textheight]{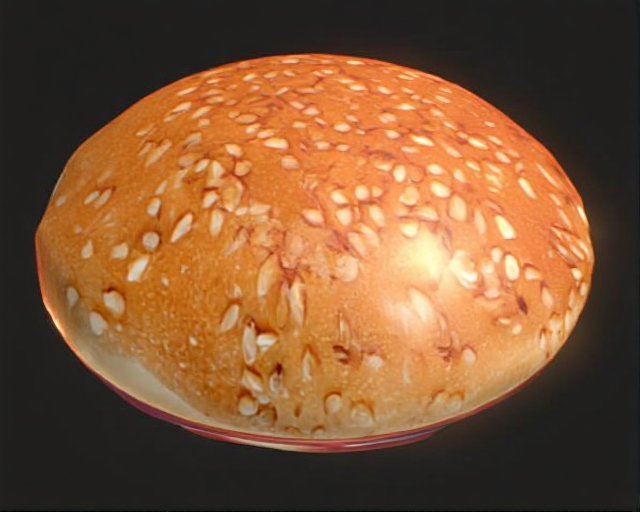} \hspace{-5mm}  
\\
\quad Input & \quad Vanilla & \quad FastV~\cite{chen2024image} & \quad PDrop~\cite{xing2025conical} & \quad IVC-Prune~\cite{sun2026ivc} & \quad \textbf{G$^2$TR (ours)} 

\end{tabular}
}
\caption{More visualization results. Instructions from top to down are: ``What does the item in the box look like when taken out?''; ``What is the appearance of these pastries after they have been baked?''; ``The fruit in the image is ready for harvest. Could you display how it looks before it is ready?''; ``Place Christian classic symbol items on the table.''; ``Draw the top view of the object based on its front view.''. Our method still indicates strong performances.}
\label{fig:edit_visual_more}
\end{figure}

We put more visualizations of image editing results in Figure~\ref{fig:edit_visual_more}. Images and instructions are from IntelligentBench~\cite{deng2025bagel} and KRIS-Bench~\cite{wukris}. Our method still keeps necessary visual details in the edited images, and follow the instructions well, better than compared baselines.

\section{Special Note}
\label{sec:special}
\looseness=-1 This paper was finished during a very difficult time. My mother was unwell and I cannot be with her. Thank my mother three thousand for bringing me up and teaching me many things.



\end{document}